%% file: main.tex
\documentclass[]{template}

\usepackage[numbers,sort&compress]{natbib}

\usepackage[utf8]{inputenc}             
\usepackage[T1]{fontenc}                
\usepackage{lmodern}                   
\usepackage{url}                        
\usepackage{booktabs}                   
\usepackage{multirow}
\usepackage{colortbl}
\usepackage{multicol}
\usepackage{amsfonts}                   
\usepackage{amssymb}                    
\usepackage{nicefrac}                   
\usepackage{microtype}                  
\usepackage[dvipsnames]{xcolor}         

\usepackage{latexsym}

\usepackage{graphicx}
\usepackage{float}
\usepackage{subcaption}
\usepackage{wrapfig}
\usepackage{lipsum}

\usepackage{bm}

\usepackage{tabularx} 
\usepackage{ragged2e} 
\newcolumntype{L}{>{\RaggedRight\hangafter=1\hangindent=0em}X}

\usepackage{enumitem}

\usepackage{amsmath}
\usepackage{amssymb}
\usepackage{mathtools}
\usepackage{amsthm}

\setboolean{logo}{true}    

\usepackage[linesnumbered,ruled,vlined]{algorithm2e}

\hypersetup{
    colorlinks=true,
    linkcolor=red,
    citecolor=Cerulean,
    filecolor=magenta,      
    urlcolor=magenta,
}

\usepackage[capitalize,noabbrev]{cleveref}
\crefname{section}{§}{§§}
\Crefname{section}{§}{§§}

\usepackage{calligra}
\DeclareMathAlphabet{\mathcalligra}{T1}{calligra}{m}{n}

\usepackage{pifont}

\theoremstyle{plain}

\theoremstyle{definition}

\theoremstyle{remark}

\renewcommand{\paragraph}[1]{\vspace{1mm}\noindent\textbf{#1}}

\DeclareCaptionLabelFormat{cont}{#1~#2\alph{ContinuedFloat}}
\usepackage{xcolor}
\definecolor{conceptblue}{RGB}{25, 85, 155}
\newcommand{\concept}[1]{\textcolor{conceptblue}{\emph{#1}}}

\usepackage[most]{tcolorbox}
\tcbset{
  promptbox/.style={
    top=10pt,
    colback=lightgray!20,
    colframe=Black,
    colbacktitle=NavyBlue,
    enhanced,
    center,
    attach boxed title to top center={yshift=-0.1in,xshift=0.0in},
    boxed title style={boxrule=0pt,colframe=white,},
  }
}

\newtcolorbox{takeawaybox}[2][]{takeawaybox, title=#2,#1}
\tcbset{
  observationbox/.style={
    top=10pt,
    colback=lightgray!20,
    colframe=Black,
    colbacktitle=YellowGreen,
    enhanced,
    center,
    attach boxed title to top center={yshift=-0.1in,xshift=0.0in},
    boxed title style={boxrule=0pt,colframe=white,},
  }
}
\newtcolorbox{observationbox}[2][]{observationbox, title=#2,#1}

\usepackage{xspace}

\newcommand\blfootnote[1]{%
  \begingroup
  \renewcommand\thefootnote{}\footnote{#1}%
  \addtocounter{footnote}{-1}%
  \endgroup
}

\usepackage{CJK}

\newtcolorbox{promptbox}[1][]{
  enhanced, breakable,
  colback=black!3, colframe=black!35,
  boxrule=0.5pt, arc=1.5pt,
  left=6pt, right=6pt, top=4pt, bottom=4pt,
  before skip=5pt, after skip=5pt,
  fontupper=\ttfamily\small,
  fonttitle=\small\bfseries, coltitle=black,
  colbacktitle=black!10, toptitle=2pt, bottomtitle=2pt,
  #1
}
\newtcolorbox{hintbox}[1][]{
  enhanced, breakable,
  colback=blue!2!white, colframe=black!25,
  boxrule=0.5pt, arc=1.5pt,
  left=6pt, right=6pt, top=4pt, bottom=4pt,
  before skip=5pt, after skip=5pt,
  fontupper=\small,
  fonttitle=\small\bfseries, coltitle=black,
  colbacktitle=blue!6!black!6, toptitle=2pt, bottomtitle=2pt,
  #1
}

\definecolor{hlbest}{RGB}{222,233,248}   
\definecolor{hlsecond}{RGB}{240,240,240} 
\newcommand{\best}[1]{\cellcolor{hlbest}\textbf{#1}}
\newcommand{\second}[1]{\cellcolor{hlsecond}\underline{#1}}

\usepackage{textcomp}
\definecolor{cgain}{RGB}{0,110,60}
\definecolor{closs}{RGB}{176,48,48}
\newcommand{\gain}[1]{\textcolor{cgain}{#1}}
\newcommand{\loss}[1]{\textcolor{closs}{#1}}
\newcommand{\mn}{\textminus}
\newcommand{\sd}[1]{{\footnotesize\textcolor{black!55}{\textpm\,#1}}}
\newcommand{\wins}[1]{{\footnotesize\textcolor{black!55}{(#1/3)}}}
\newcommand{\nil}{\textcolor{black!45}{--}}
\newcommand{\refc}{{\footnotesize\textcolor{black!45}{ref.}}}

\definecolor{lvla}{RGB}{17,30,51}
\definecolor{lvlb}{RGB}{29,58,96}
\definecolor{lvlc}{RGB}{44,88,134}
\definecolor{lvld}{RGB}{62,113,160}
\definecolor{lvle}{RGB}{110,110,110}
\DeclareRobustCommand{\LI}{\textcolor{lvla}{\textbf{L1}}}
\DeclareRobustCommand{\LII}{\textcolor{lvlb}{\textbf{L2}}}
\DeclareRobustCommand{\LIII}{\textcolor{lvlc}{\textbf{L3}}}
\DeclareRobustCommand{\LIV}{\textcolor{lvld}{\textbf{L4}}}
\DeclareRobustCommand{\LV}{\textcolor{lvle}{\textbf{\textit{L5}}}}

\newcommand{\opsd}{OPSD}

\title{What Should a Self-Teacher See? Privileged Context Design for On-Policy Self-Distillation}

\author[1,2]{Kanghui Tian}
\author[3]{Siyuan Liu}
\author[2]{Tianxiang Jiang}
\author[1]{Shuai Dong}
\author[4]{Yizhuo Li}
\author[5]{Tian Ding}
\author[6]{Yuan Guo}
\author[1]{Songze Li}
\author[4]{Haowen Hou}
\author[7]{Congcong Wang}
\author[2]{Yi Wang\textsuperscript{$\dagger$}}

\affil[1]{Fudan University}
\affil[2]{Shanghai Artificial Intelligence Laboratory}
\affil[3]{Nanjing University}
\affil[4]{Shanghai Jiao Tong University}
\affil[5]{Peking University}
\affil[6]{University of California, Los Angeles}
\affil[7]{Tongji University}

\input{sections/0.abstract}

\begin{document}

\blfootnote{$\dagger$ Corresponding author. This work was performed at Shanghai AI Laboratory.}
\blfootnote{$*$ Code is available at \url{https://github.com/tiankanghui/Self-Teacher-Context-Design}.}

\maketitle

\input{sections/1.introduction}
\input{sections/2.related}
\input{sections/3.methodology}
\input{sections/4.experiment}
\input{sections/5.conclusion}

\section*{Acknowledgments}
This work is supported by the College of Computer Science and Artificial Intelligence, Fudan University and the Shanghai Artificial Intelligence Laboratory.


\bibliographystyle{unsrtnat}
\bibliography{refs}


\clearpage
\appendix
\input{sections/appendix}



\end{document}

%% file: sections/0.abstract.tex
\begin{abstract}
More privileged information does not always make a better teacher. We study this tension in on-policy self-distillation (\opsd{}), where a self-teacher scores the student's own rollouts under privileged context, conventionally a complete reference solution that bundles the final answer with one particular reasoning path. Holding the student view and training fixed within each scale, we compare that default against three abstractions compiled offline, a \concept{named strategy}, a \concept{method-independent framing}, and a \concept{problem category}, and against an \concept{answer-only} control that keeps the destination but removes the path. In the primary runs on competition mathematics, the best intermediate contexts improve the in-domain peak mean over the full solution by 1.4 points at 4B and 1.6 at 8B, while storing an order of magnitude fewer hint tokens. Comparisons across three seeds also show positive mean gains for the framing and category contexts at both scales. Answer-only conditioning remains competitive in the primary runs, within 0.2 points of the full solution at these scales. The preferred context varies with student scale and task. Initial teacher--student KL does not order downstream performance. What a self-teacher should see is therefore not everything it could, but the level of abstraction its student can still act on.

\end{abstract}

%% file: sections/1.introduction.tex
\section{Introduction}
On-policy self-distillation (\opsd{}) promises dense reasoning supervision without a more capable online teacher. The student samples a trajectory from the problem alone, and a frozen copy of its initial checkpoint provides next-token distributions at the same prefixes while observing teacher-only information, such as a verified solution~\citep{zhao2026self}. Whereas conventional on-policy distillation typically derives its teacher advantage from greater model capability~\citep{agarwal2024onpolicy,lu2025onpolicy-thinkinglab}, \opsd{} relies on \emph{information asymmetry} instead, which makes the design of this privileged context part of the supervision mechanism.

More detailed context, however, need not provide more useful supervision: under peak reporting on competition mathematics, replacing the full solution with its final answer alone changes 4B and 8B aggregate scores by less than 0.2 points. A full worked solution supplies an explicit procedure but may anchor the teacher's guidance to a particular derivation, whereas an abstract hint or bare answer offers less procedural support and more latitude in how the problem is solved. These choices can change which continuations the teacher favors, not merely how strongly its distribution differs from the student's. This raises a central question: \emph{at what level of abstraction should a self-teacher see a reference solution, and how should that choice depend on the student and task?}

Prior work demonstrates the value of privileged solutions for self-distillation~\citep{zhao2026self}, while also identifying risks of reduced exploration and reference-path bias~\citep{kaur2026rethinking,harne2026privileged}. ATESD varies the visible reference prefix while retaining the answer~\citep{han2026adaptive}, and a broader reference-type comparison includes abstract hints~\citep{shrestha2026rethinking}. Using a training-free gradient-alignment proxy, \citet{armandpour2026unmasking} report no universally best distillation context and rank summarized demonstrations above raw ones for their larger student. What remains unclear is how adjacent semantic abstractions without explicit solution steps affect student performance across scales and tasks.

\begin{figure*}[!t]
\centering
\includegraphics[width=0.98\textwidth]{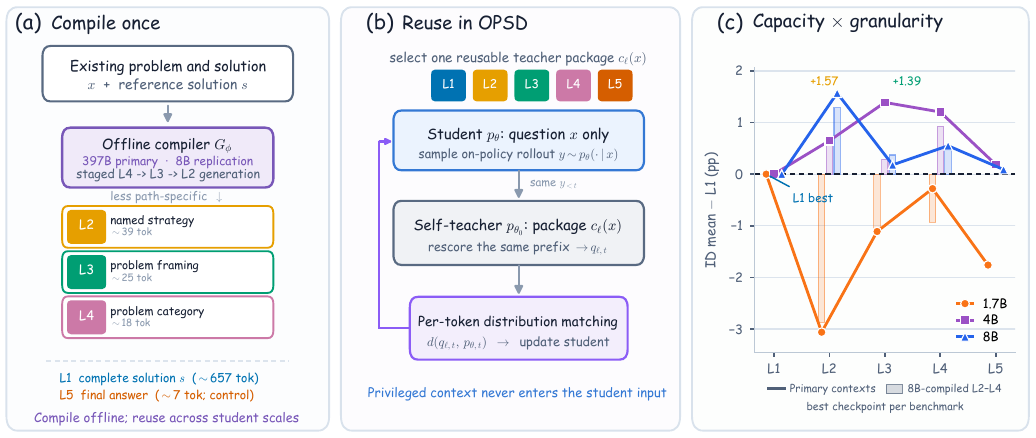}
\caption{
Study overview.
\textbf{(a)} An offline compiler produces reusable contexts from coarse to fine (L4$\to$L3$\to$L2); L1 is the original solution and L5 its final answer. Labels give mean token counts.
\textbf{(b)} The student samples $y$ from $x$ alone, while a frozen copy of the base model scores the same prefixes with additional context $c_\ell(x)$; only the student is updated.
\textbf{(c)} Peak aggregate scores relative to L1 for the primary compiler (solid lines, seed 42) and the single-run Qwen3-8B replication (pale bars). The two compilers agree at the family level but not on the best intermediate level. Replications of L1, L3, and L4 across three seeds appear in Table~\ref{tab:aggregation-summary}.
}
\label{fig:study-overview}
\end{figure*}

We study this question through controlled interventions on the context supplied to a fixed self-teacher (Figure~\ref{fig:study-overview}). Starting from the same problem--solution pairs, an offline compiler constructs three progressively more abstract context packages: a named strategy (L2), a method-independent framing (L3), and a problem category (L4), all generated under contracts that prohibit the final answer. We compare these with the full solution (L1), an answer-only condition (L5), and a control that shortens the reference to a fixed 10\% prefix plus the answer without semantic rewriting. Across three Qwen3 student scales~\citep{qwen2025qwen3}, we hold the student prompt, frozen teacher checkpoint, training set, rollout procedure, and loss fixed within each scale, and we evaluate on three competition mathematics benchmarks and four transfer benchmarks. We assess context utility through student performance after training and use initial teacher--student KL as a separate diagnostic.

Our contributions are:
\begin{itemize}
    \item \textbf{Concise semantic contexts can outperform full solutions.}
    In the primary runs, the best intermediate context exceeds the full solution by 1.4 and 1.6 points in the 4B and 8B in-domain peak means while storing an order of magnitude fewer hint tokens (Table~\ref{tab:id-main}). Across three seeds, L3 and L4 show positive mean gains over L1 at 4B and 8B under all four reporting rules. Recompiling the contexts with Qwen3-8B also preserves the peak advantage at the larger scales. The preferred representation varies with student scale and task.

    \item \textbf{Answer-only context remains competitive at the larger scales.}
    Answer-only conditioning comes within 0.2 points of the full solution's in-domain peak mean at 4B and 8B, though it trails by 1.76 points at 1.7B.

    \item \textbf{Distributional displacement does not order teaching utility.}
    The full solution induces the largest initial teacher--student KL at every scale, yet does not yield the highest in-domain peak mean at 4B or 8B. Item-level analysis shows context redistributing which problems the trained student solves.
    Privileged context is therefore best judged by the students it trains.
\end{itemize}

%% file: sections/2.related.tex
\section{Related Work}

\paragraph{On-Policy Distillation.} Knowledge distillation transfers model predictions or reasoning traces~\citep{hinton2015distilling,gu2024minillm,hsieh2023distilling}, and bootstrapped methods learn from the model's own successful rationales~\citep{zelikman2022star}.  On-policy distillation instead supplies token-level teacher feedback on trajectories sampled from the student's current policy~\citep{agarwal2024onpolicy,lu2025onpolicy-thinkinglab,exopd,uniopd}, with later work refining the objective~\citep{li2026rethinkingopd,jin2026entropyaware,luo2026demystifying}, conditioning the self-teacher on demonstrations~\citep{sdft}, or combining it with an external teacher~\citep{dopd}. Our baseline is OPSD, which conditions the frozen self-teacher on a complete reference solution~\citep{zhao2026self}; that choice of context is the variable we study.

\paragraph{Privileged Information and Reasoning Guidance.} Privileged context gives the teacher access to information withheld from the student. This makes \opsd{} an instance of learning using privileged information: the teacher sees training-time signals the student will not receive at inference~\citep{vapnik2009new,lopezpaz2016unifying}. Context distillation internalizes prompt-induced behavior into parameters~\citep{snell2022context}.  The privileged signal takes many forms: feedback and self-generated traces~\citep{sdpo,rlsd}, contextual instructions such as concision~\citep{sang2026reasoningcompression,ye2026onpolicycontext}, recoverable visual cues and other forms of privileged information~\citep{yuan2026visionopdlearningfinedetails,vicur,penaloza2026privileged}, and majority-vote pseudo-solutions requiring no external supervision~\citep{li2026uopsd}. It also enters at different stages, through partial references during generation~\citep{wu2026regft} or step-level process supervision~\citep{lightman2023verify}. Given a solution, designing the teacher context requires deciding which parts to expose and in what form.

\paragraph{Reference Exposure.} ATESD learns how much of a reference reasoning prefix to expose while
retaining the final answer~\citep{han2026adaptive}. 
Complementary analyses examine the supervision induced by reference conditioning, identifying confidence miscalibration~\citep{zhang2026illusion}, failures under long generation budgets~\citep{kaur2026rethinking}, and per-token bias toward the supplied solution~\citep{harne2026privileged}.  Reference controls distinguish problem-specific information from context-induced teacher behavior~\citep{ichihara2026privileged}.  Beyond exposure length, \citet{shrestha2026rethinking} compare teacher modes and reference types, including a method-level abstract-hint condition, and analyze how teacher supervision aligns with changes in student predictions. Using a training-free gradient-alignment proxy, \citet{armandpour2026unmasking} likewise report no universally best context, ranking summarized demonstrations above raw ones for their larger student. PS-OPSD recasts the reference as a problem-solving structure with a selected path of state transitions~\citep{zhao2026problemspace}. We instead compare three adjacent abstraction levels that omit explicit solution steps, training one student per level.

%% file: sections/3.methodology.tex
\section{Privileged Context as a Controlled Intervention}
\label{sec:method}

We treat the context package supplied to the teacher as the intervention and evaluate its utility after student training. This section formalizes the comparison, describes on-policy supervision and staged context construction, and specifies the diagnostic tests and controls.

\subsection{Problem formulation}

Let $\mathcal{D}=\{(x_i,s_i)\}_{i=1}^{N}$ be a training set of problems and their reference solutions. A context policy $c_\ell$ maps each training problem to a precomputed prompt for the teacher:
\begin{equation}
c_\ell(x)=B_\ell\!\left(x,h_\ell(x)\right),
\end{equation}
where $h_\ell(x)$ is a representation constructed from the problem and its reference solution, and $B_\ell$ is the prompt template that combines the original problem, the representation, and the wording that introduces it. We refer to this introductory wording as the \emph{bridge}. The student receives $x$ alone; the context policy affects only the supervision supplied by a frozen teacher.

In on-policy supervision, the teacher scores the student's own prefixes, which may depart from the reference derivation. We therefore test which representation of that reference provides the most useful supervision.

Our target is the utility of a context \emph{after training}, rather than a property of the teacher distribution alone. Let $\theta_{\ell,\xi,k}$ denote the student checkpoint at step $k\in\mathcal{S}$, trained from the shared initialization $\theta_0$ under context $c_\ell$ and training randomness $\xi$, with the remaining pipeline fixed; $\theta_0$ also serves as the frozen teacher checkpoint. For a benchmark $\mathcal{B}$ and a specified reporting rule $r$, we define
\begin{equation}
U_{\mathcal{B},r}(c_\ell;\theta_0)
=
\mathbb{E}_{\xi}\!\left[
r\!\left(
\{\operatorname{Score}_{\mathcal{B}}
(\theta_{\ell,\xi,k})\}_{k\in\mathcal{S}}
\right)
\right].
\label{eq:context-utility}
\end{equation}
Here, $r$ may report a peak score, a fixed endpoint, or another checkpoint aggregate. Dependence on the fixed training set and pipeline is suppressed. Context design then amounts to maximizing $U_{\mathcal{B},r}$ over a candidate set of context policies. In the primary study, each training run uses a fixed context level. We vary this level across runs to test whether the best representation changes with student scale, evaluation task, or reporting rule, instead of assuming a universally optimal context. Supplementary experiments examine heterogeneous assignment and learned routing (Section~\ref{sec:mixtures}).

\subsection{How context enters on-policy self-distillation}

Context changes the teacher's supervision targets without changing its parameters. At each update, the current student samples a trajectory $y=(y_1,\ldots,y_T)\sim p_\theta(\cdot\mid x)$, and a frozen copy of the initial checkpoint scores the same trajectory prefixes with privileged context:
\begin{equation}
p_{\theta,t}(\cdot)
=
p_\theta(\cdot\mid x,y_{<t}),
\qquad
q_{\ell,t}(\cdot)
=
p_{\theta_0}(\cdot\mid c_\ell(x),y_{<t}).
\end{equation}
At a fixed problem and prefix, changing $c_\ell(x)$ changes the effective teacher distribution even though $\theta_0$ is unchanged.

Every condition uses the same clipped full-vocabulary objective. Writing $d$ for this shared objective, the loss for each trajectory is
\begin{equation}
\mathcal{L}_{\ell}(\theta;x,y)
=
\frac{1}{T}
\sum_{t=1}^{T}
d\!\left(q_{\ell,t},p_{\theta,t}\right).
\label{eq:main-loss}
\end{equation}
Appendix~\ref{app:training-config} gives the exact clipping and reduction. The teacher is frozen and the sampled trajectory is fixed when differentiating this loss, so a context change can alter both the magnitude and the direction of $-\nabla_\theta\mathcal{L}_\ell$. Greater teacher--student disagreement need not imply a more useful update.

The context package is used only for teacher scoring: the student never receives the hint or bridge, during rollout generation or at evaluation. Within each primary comparison we hold the student initialization, student prompt, frozen teacher checkpoint, training set, rollout procedure, training budget, and loss fixed. What is fixed is the \emph{rollout procedure}, not the realized trajectories: as students learn under different contexts, their subsequent on-policy trajectories can diverge, and this feedback is part of the end-to-end effect being evaluated. Following \citet{zhao2026self}, the teacher scores with Qwen3's thinking template while student rollouts use the non-thinking template (Appendix~\ref{app:training-config}), an interface difference common to L1--L5.

\subsection{Constructing matched semantic contexts}

We compare representations that vary in their intended specificity to the reference reasoning path (Figure~\ref{fig:study-overview}(a)). L1 provides the complete reference solution. L2 names a method, theorem, or tool without working through solution steps; L3 gives a method-independent framing of the problem; L4 gives only its mathematical domain and object category. All three contracts prohibit the final answer. L5 supplies only the final answer as reference content and serves as the answer-only control.

We define the L1--L4 ordering through the semantic contracts used to generate L2--L4 (Appendix~\ref{app:generation-prompts}), and audit compliance in the resulting contexts (Table~\ref{tab:hint-audit}). Each of L2--L4 is newly written rather than a truncation of L1, so the ordering expresses intended abstraction rather than nested text, and our comparisons treat each representation as a complete context package.

\paragraph{Staged Semantic Compilation.}
Let $s(x)$ denote the existing reference solution and $G_\phi$ an offline compiler. We construct L2--L4 by refining from coarse to fine, each call receiving the problem, the reference solution, and every coarser representation already produced:
\begin{equation}
h_\ell(x)
=
G_\phi^{(\ell)}\!\left(x,s(x),\{h_{\ell'}(x)\}_{\ell<\ell'\leq4}\right),
\qquad
\ell=4,3,2,
\end{equation}
so that $h_4$ is compiled from the problem and solution alone. Passing the coarser outputs to finer calls makes the intended semantic boundaries explicit. L1 uses the original solution and L5 extracts its final answer, so neither requires compilation; Appendix~\ref{app:hint-example} shows all conditions for one problem.

The primary contexts are compiled with Qwen3.5-397B-A17B~\citep{qwen3.5}, and we repeat the same staged contracts with Qwen3-8B in thinking mode to test sensitivity to the choice of compiler. Both compilers receive the existing reference solutions, and all contexts are precomputed, so neither compiler is queried during \opsd{} training.

\subsection{Utility estimation and diagnostic tests}

We evaluate context utility through student scores after training under the reporting rule in Equation~\ref{eq:context-utility}. For the in-domain comparison, we train students under L1, L3, and L4 with three seeds at each scale and apply four checkpoint aggregation rules, two of them selection-free. For L2 and L5, the transfer evaluations, and the supplementary controls, we report one training run per condition rather than averages across seeds. Where reported, resampling matched items quantifies evaluation uncertainty, not variability across training runs. Peak and selection-free rules define different estimands and are distinguished throughout.

\paragraph{Distributional Discrepancy.}
We separately measure how much a context changes the frozen base model's next-token distribution. Within each scale, let $\mu_0$
denote the shared empirical distribution of problem--prefix pairs from fixed rollouts of the base model. We compute
\begin{equation}
D(c_\ell)
=
\mathbb{E}_{(x,y_{<t})\sim\mu_0}
\left[
D_{\mathrm{KL}}\!\left(
q_{\ell,t}\,\Vert\,p_{\theta_0,t}
\right)
\right],
\label{eq:context-displacement}
\end{equation}
where
$p_{\theta_0,t}(\cdot)=p_{\theta_0}(\cdot\mid x,y_{<t})$. Using the same prefixes from the base model across contexts avoids mixing changes induced by context with differences in trained students' rollout distributions. This diagnostic measures distributional discrepancy; it is not the clipped training objective, an update norm, or a direct measure of teacher quality.

\paragraph{Empirical Predictions.}
We test three deliberately strong baseline predictions, none of which is assumed by the formulation. \emph{Full-solution dominance} predicts that L1 yields the greatest downstream utility. \emph{Discrepancy-based ordering} predicts that contexts with larger $D(c_\ell)$ also yield greater utility at a fixed student scale, task, and reporting rule. \emph{Context invariance} predicts that one context is preferred across student scales, tasks, and reporting rules; we test it through ranking changes across scales and benchmarks, with item-level gains and losses relative to L1 describing how a given ranking arises.

\subsection{Intervention validity and scope}
\label{sec:validity}

In the primary comparisons, we vary the complete context package while holding the source problem--solution pairs and the training recipe fixed within each scale, so context is the only variable. The packages can still differ in semantic content, length, answer access, and, in the primary runs, bridge wording. Hints generated from the same reference solutions can also differ in quality. These comparisons therefore evaluate the packages as a whole rather than isolating the causal effect of semantic abstraction.

L1 versus L2--L4 tests whether exposing the complete derivation is preferable to the intermediate packages. Comparisons within L2--L4 test the utility of adjacent semantic roles, and L5 tests supervision from the final answer alone.

Across scales the student and training recipe change together, including the effective batch size (Table~\ref{tab:training-config}), which makes the scale comparison one between trained systems rather than parameter count alone. Section~\ref{sec:controls} reports controls that each narrow a specific alternative explanation.

\paragraph{Context Length.}
We distinguish the size of the reusable hint from the length of the teacher's context prompt. Under the Qwen3 tokenizer, the primary L2--L4 hints average 38.9, 25.2, and 17.7 tokens against 657.3 for L1, a 16.9--37.1$\times$ reduction in stored hint content; including the problem and bridge, the context prompts average 225.5, 212.7, and 203.3 tokens against 835.0, a 3.7--4.1$\times$ reduction (Appendix Figure~\ref{fig:intervention-audit}(a); Table~\ref{tab:hint-lengths}). Both quantify context cost rather than semantic content.

\paragraph{Content Audits.}
Full-corpus scans find no empty hints, generation-error sentinels, or lexical answer matches for either compiler. A matched audit of 1,200 problems finds no direct answer copying, but does detect answer-equivalent leakage: 1.42\% at L2 and 0.33\% at L3 for the primary compiler, rising to 4.17\% and 2.58\% for Qwen3-8B, with L4 at 0.08\% and 0.17\% (Table~\ref{tab:hint-audit}). Appendix~\ref{app:audits} gives the definitions and prompts for this model-assisted audit.

%% file: sections/4.experiment.tex
\section{Experiments}
\label{sec:experiments}

We train Qwen3-1.7B, 4B, and 8B~\citep{qwen2025qwen3} for 200 steps with LoRA~\citep{hu2021loralowrankadaptationlarge} on the 29,434-problem \opsd{} mathematics pool~\citep{zhao2026self,guha2025openthoughtsdatarecipesreasoning}, checkpointing every 25 steps. At each scale, the student and the frozen self-teacher start from the same checkpoint, and the primary L1--L5 runs share the data order, rollout procedure, objective, and recipe, with no tuning by context level (Appendices~\ref{app:training-config} and~\ref{app:bridge-prompts}). Base denotes the unadapted checkpoint.

Avg@12 averages correctness over twelve sampled completions per problem. Following the original \opsd{} convention, we report each in-domain (ID) benchmark's best Avg@12 across the eight checkpoints. The \emph{ID peak mean} averages these separately selected peaks on AIME24, AIME25, and HMMT25, so it need not represent a single checkpoint; we also report the selection-free step-200 mean and every checkpoint (Appendix~\ref{app:training-curves}). At step 200, transfer evaluation uses MT-AIME2024-7Lang~\citep{son2025linguistic} (macro Avg@12 over seven languages), AutoLogi-EN~\citep{zhu2025autologi} (verifier accuracy), GPQA-Diamond~\citep{rein2023gpqa} (Avg@10), and ZebraLogic-grid~\citep{lin2025zebralogic} (exact puzzle accuracy; Appendix~\ref{app:setup-details}). All score differences are in percentage points.

\subsection{Context utility varies with student scale and task}
\label{sec:results}

\paragraph{In-domain Performance.}
The best intermediate context exceeds the full solution by 1.39 points at 4B and 1.57 at 8B, while L1 keeps the highest peak mean at 1.7B, 0.28 ahead of L4 (Table~\ref{tab:id-main}). The winning level does not shift monotonically toward coarser context with scale. At step 200, L1 still leads at 1.7B, while at 4B both L2 and L4 reach 62.59 against 60.93 for L1, and at 8B L3 reaches 65.37 against 65.28 (Appendix~\ref{app:training-curves}).

\begin{table*}[t]
\caption{In-domain peak Avg@12 (\%), primary compiler for L2--L4. All entries are the primary run (seed 42); Table~\ref{tab:aggregation-summary} reports replication of \LI, \LIII, and \LIV{} across three seeds. Mean averages separately selected benchmark peaks; Base is evaluated once. Bold marks the highest score among the trained conditions in each row. The \emph{vs.\ \LI} rows give each column's mean minus the mean for the full solution in percentage points, \gain{green} above \LI{} and \loss{red} below. Selected steps are listed in Appendix~\ref{app:training-curves}.}
\label{tab:id-main}
\centering
\small
\setlength{\tabcolsep}{4pt}
\begin{tabular*}{\textwidth}{@{\extracolsep{\fill}}llcccccc@{}}
\toprule
Model & Benchmark & Base & \LI: solution & \LII: strategy & \LIII: framing & \LIV: category & \LV: answer \\
\midrule
\multirow{4}{*}{1.7B}
 & AIME24 & 49.72 & 57.22 & 56.11 & 56.94 & \textbf{59.72} & 57.78 \\
 & AIME25 & 36.39 & \textbf{43.33} & 38.33 & 41.39 & 43.06 & 41.67 \\
 & HMMT25 & 21.67 & \textbf{31.11} & 28.06 & 30.00 & 28.06 & 26.94 \\
\cmidrule(lr){2-8}
 & Mean & 35.93 & \textbf{43.89} & 40.83 & 42.78 & 43.61 & 42.13 \\
 & \emph{vs.\ \LI} & \loss{\mn7.96} & \refc & \loss{\mn3.06} & \loss{\mn1.11} & \loss{\mn0.28} & \loss{\mn1.76} \\
\midrule
\multirow{4}{*}{4B}
 & AIME24 & 75.00 & 76.94 & 76.94 & \textbf{77.50} & 76.94 & 75.28 \\
 & AIME25 & 64.44 & 68.61 & 69.44 & 69.72 & 69.17 & \textbf{70.56} \\
 & HMMT25 & 41.39 & 44.72 & 45.83 & 47.22 & \textbf{47.78} & 45.00 \\
\cmidrule(lr){2-8}
 & Mean & 60.28 & 63.43 & 64.07 & \textbf{64.81} & 64.63 & 63.61 \\
 & \emph{vs.\ \LI} & \loss{\mn3.15} & \refc & \gain{+0.65} & \gain{+1.39} & \gain{+1.20} & \gain{+0.19} \\
\midrule
\multirow{4}{*}{8B}
 & AIME24 & 75.28 & 78.89 & 79.72 & 79.44 & \textbf{80.56} & 78.61 \\
 & AIME25 & 66.94 & 73.06 & \textbf{74.17} & 71.39 & 71.39 & 72.78 \\
 & HMMT25 & 44.17 & 48.06 & \textbf{50.83} & 49.72 & 49.72 & 48.89 \\
\cmidrule(lr){2-8}
 & Mean & 62.13 & 66.67 & \textbf{68.24} & 66.85 & 67.22 & 66.76 \\
 & \emph{vs.\ \LI} & \loss{\mn4.54} & \refc & \gain{+1.57} & \gain{+0.19} & \gain{+0.56} & \gain{+0.09} \\
\bottomrule
\end{tabular*}
\end{table*}

\begin{table}[t]
\caption{No context leads across the whole transfer suite. For each scale and benchmark, \emph{Winner} is the highest-scoring context at step 200 (primary runs, seed 42) and $\Delta$ its margin over L1 in percentage points; \nil{} marks the two cells where L1 itself wins. The last row counts how many distinct contexts win the four benchmarks at that scale. Full scores are in Appendix Table~\ref{tab:ood-main}.}
\label{tab:transfer-summary}
\centering
\small
\setlength{\tabcolsep}{4pt}
\begin{tabular*}{\linewidth}{@{\extracolsep{\fill}}lcrcrcr@{}}
\toprule
& \multicolumn{2}{c}{1.7B} & \multicolumn{2}{c}{4B} & \multicolumn{2}{c}{8B} \\
\cmidrule(lr){2-3} \cmidrule(lr){4-5} \cmidrule(lr){6-7}
Benchmark & Winner & $\Delta$ & Winner & $\Delta$ & Winner & $\Delta$ \\
\midrule
MT-AIME-7Lang & \LIII & \gain{+1.71} & \LIV & \gain{+2.10} & \LII & \gain{+1.71} \\
GPQA-Diamond  & \LIII & \gain{+0.81} & \LII & \gain{+1.21} & \LIII & \gain{+1.41} \\
ZebraLogic    & \LI & \nil & \LIV & \gain{+0.40} & \LIII & \gain{+1.70} \\
AutoLogi      & \LIV & \gain{+0.19} & \LV & \gain{+0.38} & \LI & \nil \\
\midrule
Distinct winners & \multicolumn{2}{c}{3} & \multicolumn{2}{c}{3} & \multicolumn{2}{c}{3} \\
\bottomrule
\end{tabular*}
\end{table}

\paragraph{Transfer Rankings Vary by Task.}
At step 200, different transfer tasks favor different contexts (Table~\ref{tab:transfer-summary}), and this variation appears even within a single scale: at 1.7B, L3 leads on MT-AIME and GPQA while L1 leads on ZebraLogic. Because all conditions are evaluated at step 200, these ranking changes do not result from selecting different checkpoints for different conditions.

\paragraph{Answer-only Context.}
\label{sec:answer-ablation}
L5 is within 0.2 points of L1's ID peak mean at 4B and 8B, but trails L1 by 1.76 points at 1.7B (Table~\ref{tab:id-main}). Answer access without supplied reasoning is therefore competitive without reaching the highest observed utility: the leading intermediate context exceeds L5 at both larger scales, and on ZebraLogic at 8B, L3 scores 88.10 against 85.90.

\subsection{Robustness and alternative explanations}
\label{sec:controls}

\paragraph{Checkpoint Aggregation and Training Seeds.}
We replicate the L1, L3, and L4 in-domain comparison with three training seeds at every scale and aggregate each run under four checkpoint rules (Table~\ref{tab:aggregation-summary}). At 4B and 8B, the mean paired gains of both L3 and L4 over L1 are positive under all four rules, and under the selection-free mean across all checkpoints every individual seed favors both intermediate levels. L4 attains a positive paired peak difference in all three seeds at both scales, whereas L3 does not at 4B seed~43 or 8B seed~44; at the endpoint L3 leads in all six runs and L4 in three. At 1.7B, the mean paired peak differences are $-0.12$ and $0.00$ points for L3 and L4, respectively; the two selection-free rules favor L1 on average. The endpoint at step 200 carries the largest paired standard deviation in five of the six contrasts. This is the pattern the remaining controls follow: they reproduce the contrast between full and intermediate contexts more consistently than any particular winner among L2--L4.

\begin{table}[t]
\caption{In-domain differences from L1, paired by training seed (42--44). Cells show mean $\pm$ sample SD in percentage points; $(n/3)$ counts strictly positive differences. SD describes variation across seeds, not a confidence interval. Peak selects each benchmark's best checkpoint; best common selects one checkpoint per run by the three-benchmark mean. Green/red indicate positive/negative mean differences. Absolute scores are in Appendix~\ref{app:multi-seed}.}
\label{tab:aggregation-summary}
\centering
\small
\setlength{\tabcolsep}{4pt}
\begin{tabular*}{\columnwidth}{@{\extracolsep{\fill}}llcccc@{}}
\toprule
& & \multicolumn{2}{c}{Checkpoint selected by score} & \multicolumn{2}{c}{Selection-free} \\
\cmidrule(lr){3-4} \cmidrule(lr){5-6}
Student & Contrast & Per-bench.\ peak & Best common & All-ckpt.\ mean & Step 200 \\
\midrule
\multirow{2}{*}{1.7B} & \LIII\,\mn\,\LI & \loss{\mn0.12}\,\sd{0.86}\,\wins{2} & \loss{\mn0.12}\,\sd{0.47}\,\wins{1} & \loss{\mn0.14}\,\sd{0.70}\,\wins{1} & \loss{\mn0.83}\,\sd{2.33}\,\wins{1} \\
 & \LIV\,\mn\,\LI & 0.00\,\sd{0.33}\,\wins{1} & \gain{+0.06}\,\sd{0.05}\,\wins{2} & \loss{\mn0.32}\,\sd{0.68}\,\wins{1} & \loss{\mn1.33}\,\sd{2.04}\,\wins{1} \\
\midrule
\multirow{2}{*}{4B} & \LIII\,\mn\,\LI & \gain{+0.71}\,\sd{0.69}\,\wins{2} & \gain{+0.90}\,\sd{0.23}\,\wins{3} & \gain{+1.03}\,\sd{0.15}\,\wins{3} & \gain{+1.67}\,\sd{0.73}\,\wins{3} \\
 & \LIV\,\mn\,\LI & \gain{+0.96}\,\sd{0.21}\,\wins{3} & \gain{+1.64}\,\sd{0.60}\,\wins{3} & \gain{+1.01}\,\sd{0.32}\,\wins{3} & \gain{+1.05}\,\sd{1.32}\,\wins{2} \\
\midrule
\multirow{2}{*}{8B} & \LIII\,\mn\,\LI & \gain{+0.25}\,\sd{0.65}\,\wins{2} & \gain{+0.62}\,\sd{0.44}\,\wins{3} & \gain{+0.49}\,\sd{0.34}\,\wins{3} & \gain{+0.28}\,\sd{0.32}\,\wins{3} \\
 & \LIV\,\mn\,\LI & \gain{+0.49}\,\sd{0.37}\,\wins{3} & \gain{+0.80}\,\sd{0.81}\,\wins{2} & \gain{+0.62}\,\sd{0.13}\,\wins{3} & \gain{+0.22}\,\sd{1.05}\,\wins{1} \\
\bottomrule
\end{tabular*}
\end{table}

\paragraph{Compiler Replication.}
Regenerating L2--L4 with Qwen3-8B, under the same semantic contracts and training conditions, preserves the ID peak ordering between full and intermediate contexts: L1 leads at 1.7B, and intermediate contexts exceed it at 4B and 8B. At 8B the compiler and student have the same nominal parameter count. The preferred intermediate changes at 4B, from L3 to L4. The 8B compiler also shows higher L2/L3 answer-equivalent leakage, so the packages are not identical in quality; its L4 leakage is only 0.17\%, however, and L4 likewise trails L1 at 1.7B and exceeds it at 4B and 8B (Appendix~\ref{app:compiler-replication}; Table~\ref{tab:hint-audit}).

\paragraph{Semantic Rewriting vs.\ Prefix Truncation.}
A control using a 10\% reference prefix plus the answer, inspired by \citet{han2026adaptive}, tests whether shortening the reference without semantic rewriting recovers the observed gains. In one 4B run, the prefix control reaches an ID peak mean of 63.89, above L1 and L5 but below all three intermediate contexts, with L3 and L4 ahead by 0.93 and 0.74 points (Appendix~\ref{app:prefix-control}, Table~\ref{tab:prefix-ablation}). At step 200 it nearly matches L3 (61.67 against 61.76), though L4 remains 0.93 points ahead. Semantic rewriting therefore yields the higher peaks here, and like the other conditions, the prefix differs from L2--L4 in answer access as well as length (Section~\ref{sec:validity}).

\paragraph{Bridge Wording and Rollout Length.}
With a bridge shared across levels, every intermediate context exceeds L1 by 1.0--2.3 ID peak points (Table~\ref{tab:shared-bridge}), and doubling the rollout budget from 1,024 to 2,048 tokens leaves every intermediate context at or above L1 on the peak mean (Appendix~\ref{app:budget-robustness}). Neither bridge wording nor the original length limit is therefore sufficient to explain the contrast.

\paragraph{Mixtures and Routing.}
\label{sec:mixtures}
Heterogeneous context assignment does not consistently improve on a fixed level. Across balanced mixtures and four routing policies, no run exceeds the best fixed-level ID peak, though some exceed the best fixed-level mean at step 200 (Appendix~\ref{app:shared-bridge}).

\subsection{Distributional discrepancy does not order context utility}
\label{sec:mechanism}

For each scale, the unadapted base model generates two rollouts on each of 600 training problems using the exact student prompt. We rescore identical completion tokens under the student prompt, the L1--L5 teacher packages, and a reference using the teacher interface without a hint, computing unclipped full-vocabulary $D_{\mathrm{KL}}(q_{\ell,t}\Vert p_{\theta_0,t})$ on shared prefixes as in Equation~\ref{eq:context-displacement} (Appendix~\ref{app:distributional}).

\begin{figure*}[t]
\centering
\includegraphics[width=0.96\textwidth]{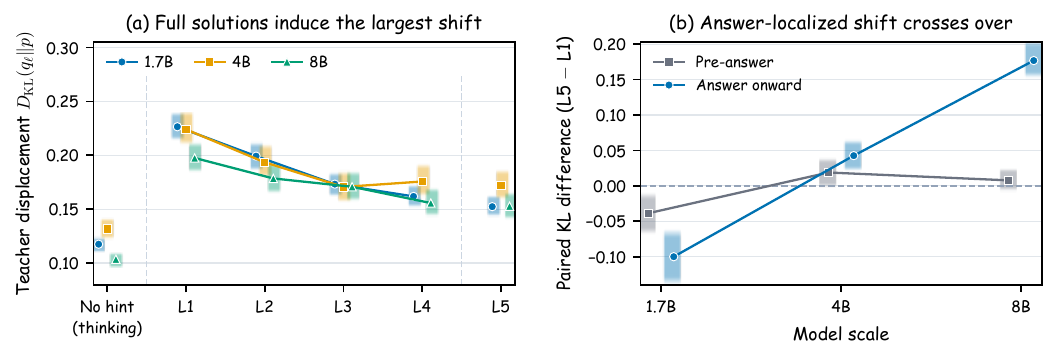}
\caption{Distributional discrepancy induced by context at initialization, measured on fixed rollouts from the base model with unclipped full-vocabulary KL (95\% bootstrap intervals clustered by problem). \textbf{(a)} Mean teacher-to-student KL per context; the no-hint interface and L5 are reference conditions outside the L1--L4 hierarchy. \textbf{(b)} Paired $\mathrm{L5}-\mathrm{L1}$ differences in pre-answer and answer-onward windows around the final \texttt{\textbackslash boxed} marker (Appendix~\ref{app:answer-aligned-geometry}).}
\label{fig:supervision-geometry}
\end{figure*}

L1 has the largest mean KL at every scale (Figure~\ref{fig:supervision-geometry}(a)). Every package displaces the base distribution well beyond the no-hint interface, so the displacement reflects the supplied context rather than the teacher template alone. Intermediate contexts nevertheless achieve higher 4B and 8B ID peak means with smaller initial discrepancies, so these KL values do not monotonically order the observed peak utilities. Training-free analysis points the same way, reporting only weak within-path correlations between teacher--student divergence and the usefulness of the resulting gradient~\citep{armandpour2026unmasking}.

\paragraph{Answer-Aligned Differences also Vary with Scale.}
These means average over all token positions, so we also examine the KL around the final answer. On the subset of rollouts aligned at the final boxed answer, the answer-onward $\mathrm{L5}-\mathrm{L1}$ KL difference is $-0.100$ at 1.7B but $+0.043$ and $+0.177$ at 4B and 8B (Figure~\ref{fig:supervision-geometry}(b)). Answer-only context therefore displaces the answer region less than the full solution at the smallest scale and more at the larger ones, as a localized distributional comparison rather than a measure of answer correctness.

\subsection{Context changes which problems are solved}
\label{sec:behavior-analysis}
\label{sec:success-redistribution}

On MT-AIME, GPQA, and ZebraLogic, we compare L1 with the highest-scoring intermediate condition at step 200 for each scale--task pair. Let $\hat{a}^{\mathrm{L1}}_i$ denote L1's fraction of correct completions on item $i$. We classify items as unsolved ($\hat{a}^{\mathrm{L1}}_i=0$), hard
($(0,0.25)$), medium ($[0.25,0.75)$), or easy ($[0.75,1]$); ZebraLogic's binary outcomes populate only the unsolved and easy buckets. The intermediate condition is selected on the same evaluation used for the comparison, so these analyses describe outcome differences rather than evaluate a selection policy.

\begin{wrapfigure}{r}{0.42\linewidth}
\vspace{-\intextsep}
\centering
\includegraphics[width=\linewidth]{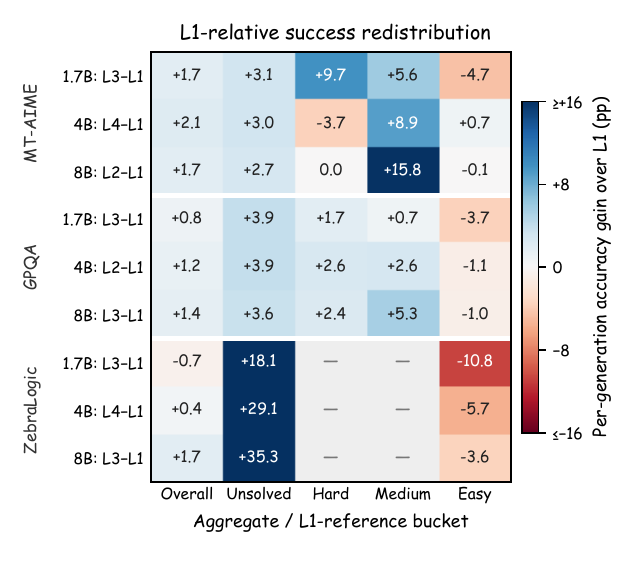}
\caption{Per-generation accuracy differences at step 200 between L1 and the highest-scoring intermediate context for each scale--task pair, by difficulty bucket under L1. Positive values favor the intermediate context.}
\label{fig:redistribution}
\end{wrapfigure}

\paragraph{Gains are Not Uniform across Difficulty Buckets.}
Intermediate contexts recover some items unsolved by L1 and improve accuracy in the medium bucket, while sometimes reducing accuracy in L1's easy bucket (Figure~\ref{fig:redistribution}; Appendix Table~\ref{tab:difficulty-redistribution}). To check dependence on using L1 to define difficulty, we also group items by Base success rates, keeping the same thresholds and the same already selected intermediate condition, so only the stratification variable changes. Across the six MT-AIME and GPQA scale--task pairs, differences in the medium bucket remain positive at 1.14--10.12 points, whereas differences in the easy bucket range from $-0.77$ to $+1.65$ points.

\paragraph{Solved-Set Differences Explain the ZebraLogic Contrast.}
The fraction of puzzles unsolved by L1 that the selected intermediate rescues rises from 18\% at 1.7B to 35\% at 8B, while losses on previously solved puzzles shrink. The net accuracy difference changes from $-0.70$ to $+1.70$ points (Appendix Figure~\ref{fig:solved-set-rotation}). At 1.7B, therefore, recovering some previously unsolved puzzles is compatible with lower overall accuracy; at 8B, the balance of gains and losses favors the intermediate context.

\paragraph{Behavioral Shifts Vary across Scales.}
We additionally compare model-judged behaviors on matched AIME24 trajectories, selecting the intermediate context by its score on the full AIME24 evaluation at step 200 and including L5 as a separate contrast. Per scale, relative to L1, backtracking rises at 1.7B and falls at 4B and 8B, while explicit correction rises at 1.7B, is unchanged at 4B, and falls at 8B (Table~\ref{tab:judge-behavior}); Figure~\ref{fig:judge-behavior} pools these scales into a single aggregate (Appendix~\ref{app:behavior-appendix}). Because these shifts differ across scales, the pooled estimate should not be read as showing behavioral invariance. Neither the per-scale nor the pooled results support an account in which intermediate contexts help by consistently increasing observable backtracking or correction.

\subsection{Interpretation and limits}
\label{sec:interpretation}

One explanation for these patterns is that full solutions and abstract contexts place different demands on the self-teacher. L1 supplies an executed procedure, which may help when the model cannot reconstruct one from a sparse cue but may also anchor its guidance to the reference derivation, whereas L2--L4 leave more of the procedure to the self-teacher when it evaluates prefixes generated by the student. Teacher and student capacity covary through the shared base checkpoint, so varying the compiler does not separate the teacher's ability to use the context from the student's ability to learn from the resulting targets. The analyses at step 200 characterize differences in item-level success and observable reasoning behavior; they do not identify what produces the separately selected peak gains (Section~\ref{sec:validity}; Appendix~\ref{app:additional-limitations}).

%% file: sections/5.conclusion.tex
\section{Conclusion}
\label{sec:discussion}

In \opsd{}, privileged context serves as a tunable supervision interface,
shaping the targets a self-teacher provides on the student's own trajectories.
What, then, should a self-teacher see? Our comparisons show that concise
semantic abstractions can outperform full solutions, with the preferred
representation depending on student scale and task. A larger initial
teacher--student KL does not guarantee better learning outcomes. Item-level
analyses further show that context choice produces gains on some problems and
losses on others. Reliably selecting context granularity from signals available
during training remains an open challenge. Through controlled comparisons of
context representations, our study establishes context granularity as a design
axis in \opsd{}. The guiding principle is to evaluate what the teacher sees by
what the student learns.

%% file: sections/appendix.tex
\section{Training and Evaluation Details}

\subsection{Training Configuration}
\label{app:training-config}

We specify the exact loss used by our trainer implementation.  At each valid
completion position, teacher and student logits are divided by the same
scoring temperature $\gamma$ and normalized over the full vocabulary:
\begin{equation}
\bar q_{\ell,t}=\operatorname{softmax}(z^q_{\ell,t}/\gamma),
\qquad
\bar p_{\theta,t}=\operatorname{softmax}(z^p_{\theta,t}/\gamma).
\end{equation}
With the configured generalized-divergence endpoint $\beta=0$
\citep{agarwal2024onpolicy}, the unreduced
vocabulary contribution is
\begin{equation}
u_{\ell,t,v}
=\bar q_{\ell,t}(v)
\left[\log \bar q_{\ell,t}(v)-\log \bar p_{\theta,t}(v)\right].
\end{equation}
Before reduction, the implementation applies an upper clamp to each element of
the resulting batch--position--vocabulary tensor:
\begin{equation}
\mathcal{L}_{\ell}(\theta)
=\frac{1}{T}\sum_{t=1}^{T}\sum_{v\in\mathcal{V}}
\min\!\left(u_{\ell,t,v},\tau\right).
\label{eq:opsd}
\end{equation}
Without the clamp this vocabulary sum would be the teacher-to-student forward
KL $D_{\mathrm{KL}}(\bar q_{\ell,t}\Vert\bar p_{\theta,t})$; with it, the
objective is a clipped teacher--student distribution objective rather than an
exact KL divergence.  Because the clamp precedes the vocabulary sum, it is a
vocabulary-component clamp, not a scalar per-token KL clip.  The inner sum is
the per-position discrepancy $d(q_{\ell,t},p_{\theta,t})$ of
Equation~\ref{eq:main-loss}, and prompt and padding positions are masked.  The clamp thresholds follow the original
\opsd{} configuration for each student scale
(Table~\ref{tab:training-config}) and are identical across context
conditions; $\tau$ is distinct from the gradient-norm cap listed in the same
table.

\begin{table}[!htbp]
\caption{Training hyperparameters.  Scale-dependent values are shown
explicitly; all remaining entries are shared.}
\label{tab:training-config}
\centering
\small
\begin{tabular}{@{}llll@{}}
\toprule
Hyperparameter & 1.7B & 4B & 8B \\
\midrule
Training examples & 29,434 & 29,434 & 29,434 \\
GPU count & 8 & 8 & 8 \\
Per-device batch & 4 & 4 & 2 \\
Gradient accumulation & 2 & 1 & 2 \\
Effective batch & 64 & 32 & 32 \\
Optimizer steps & 200 & 200 & 200 \\
Checkpoint interval & 25 & 25 & 25 \\
Learning rate & $5\times 10^{-6}$ & $5\times 10^{-6}$ & $5\times 10^{-6}$ \\
Gradient-norm cap & 0.1 & 0.1 & 0.1 \\
LoRA rank / scaling & 64 / 128 & 64 / 128 & 64 / 128 \\
Max. training rollout length & 1,024 & 1,024 & 1,024 \\
Rollout / scoring temperature & 1.1 / 1.1 & 1.1 / 1.1 & 1.1 / 1.1 \\
Scoring distribution & Full vocabulary & Full vocabulary & Full vocabulary \\
Vocabulary-component clip $\tau$ & 0.05 & 0.05 & 0.06 \\
\bottomrule
\end{tabular}
\end{table}

All scales use rollout sampling with temperature 1.1, top-$p$ 0.95, and
top-$k$ 20.  These truncation parameters affect generation only; the loss in
Equation~\ref{eq:opsd} uses the full vocabulary.  The training rollout uses
Qwen3's non-thinking chat template, while teacher scoring uses its thinking
template; evaluation enables thinking mode.  The teacher is the fixed
base model with LoRA disabled.  All runs use a single node with
8 NVIDIA H200 GPUs.

\subsection{Evaluation Protocol}
\label{app:setup-details}

Each AIME24, AIME25, and HMMT25 problem is sampled 12 times at temperature
1.0 with a maximum generation length of 38,912 tokens; Avg@$N$ denotes
accuracy averaged over $N$ sampled completions.  The main in-domain table
reports the best score observed up to step 200 separately for each (model,
condition, benchmark) triple, so an average across three benchmarks can combine
scores from different checkpoints.  Each unadapted Base is evaluated once
under identical decoding and requires no checkpoint selection.  The
peak protocol takes a maximum over eight correlated
evaluations per condition and is therefore upwardly biased relative to any
single-checkpoint score; Appendix~\ref{app:training-curves} quantifies the
gap between the best score and the score at step 200.  Differences quoted in the text are
computed from unrounded values and may deviate by up to 0.01 points from
differences recomputed from rounded table entries.

The transfer suite spans different forms of distribution shift:
\begin{itemize}
    \item \textbf{MT-AIME2024-7Lang}~\citep{son2025linguistic}: the same 30
    AIME problems translated into Chinese, French, Russian, German, Arabic,
    Japanese, and Korean.  We report the macro average of language-specific
    Avg@12 scores, weighting languages equally.
    \item \textbf{AutoLogi-EN}~\citep{zhu2025autologi}: 1,575 open-ended
    arrangement problems scored by the dataset's executable verifiers, one
    generation per problem.
    \item \textbf{GPQA-Diamond}~\citep{rein2023gpqa}: 198 graduate-level
    science questions, ten samples per question (Avg@10).
    \item \textbf{ZebraLogic-grid}~\citep{lin2025zebralogic}: 1,000
    logic-grid puzzles across 25 grid sizes.  The primary metric is exact
    puzzle accuracy; cell accuracy and difficulty buckets are secondary
    diagnostics.
\end{itemize}
C-Eval~\citep{huang2023ceval} serves as a broad Chinese knowledge audit under
a bounded generation budget; because its result is sensitive to whether a model
finishes reasoning within the output budget, it is reported separately from
the four reasoning-transfer benchmarks (Table~\ref{tab:ceval-32k}).

\section{Context Construction and Audits}

\subsection{Context Generation and a Worked Example}
\label{app:hint-example}

To make the granularity axis concrete, we show all five teacher contexts
constructed from the same competition mathematics problem.  The student sees
only the \textbf{Problem} field; during scoring, the teacher receives the
complete package $c_\ell(x)$, whose privileged content is $h_\ell(x)$.  The
boxes below show the hint content; the corresponding bridges are reproduced
in Appendix~\ref{app:bridge-prompts}.  The example is chosen from the
training pool used across all conditions.  The stored fields are reproduced
faithfully: L2--L5 are verbatim, while L1 preserves every reasoning step
with Markdown and minor grammar normalized for typesetting.  Its step~(1)
is labeled ``powers of the same base'' although the rewriting actually
unifies the exponent at $1003$; we keep the reference wording as stored,
since the conditions are compared on the packages the teacher receives.

\begin{hintbox}[title={Problem $x$}]
Let $p=2^{3009}$, $q=3^{2006}$, and $r=5^{1003}$. Which of the following
statements is true?
(A)~$p<q<r$\quad (B)~$p<r<q$\quad (C)~$q<p<r$\quad (D)~$r<p<q$\quad
(E)~$q<r<p$
\end{hintbox}

\begin{hintbox}[title={L1: complete reference solution ($h_1$)}]
Given the values: $p=2^{3009}$, $q=3^{2006}$, $r=5^{1003}$.
(1)~Express $p$ and $q$ in terms of powers of the same base:
$p=2^{3009}=2^{3\times1003}=(2^3)^{1003}=8^{1003}$ and
$q=3^{2006}=3^{2\times1003}=(3^2)^{1003}=9^{1003}$.
Note: $r=5^{1003}$ is already expressed as a power.
(2)~Compare the magnitudes: to compare $p$, $q$, and $r$, we now look at the
factors $8$, $9$, and $5$, which are $8^{1003}$, $9^{1003}$, and $5^{1003}$.
(3)~Analyze the base values: base 8 is $8=2^3$; base 9 is $9=3^2$; base 5
is~$5$.
(4)~Since the common exponent in all terms is $1003$, we compare the bases:
clearly, $5<8<9$.
(5)~Therefore, since all exponents are equal, the order of the original terms
follows the order of their bases: $5^{1003}<8^{1003}<9^{1003}$, so $r<p<q$.
Thus, the true statement is $r<p<q$, and the answer is $\boxed{\mathrm{D}}$.
\end{hintbox}

\begin{hintbox}[title={L2: named strategy ($h_2$)}]
Apply the power-of-a-power rule to rewrite each term with a common exponent
equal to the greatest common divisor of the original exponents, enabling
direct comparison of the resulting bases.
\end{hintbox}

\begin{hintbox}[title={L3: method-independent framing ($h_3$)}]
Reframe the terms into a common representational form so that comparison
depends on a single varying attribute.
\end{hintbox}

\begin{hintbox}[title={L4: problem category ($h_4$)}]
Elementary algebra: ordering of exponential terms with different bases and
exponents.
\end{hintbox}

\begin{hintbox}[title={L5: answer only ($h_5$)}]
\texttt{D}
\end{hintbox}

The intended sequence imposes progressively more restrictive roles: relative
to L1, L2 omits the executed derivation and answer while retaining a concrete
method; L3 omits method identity and retains only a framing shift; L4 retains
only the category.  Because L2--L4 are separately generated natural-language
packages, these relations are operational boundaries, not literal
set-theoretic containment.  L5 is not on this axis: it retains only the
answer.

\phantomsection
\label{app:generation-prompts}
The staged prompts that enforce these boundaries follow.
L2--L4 are generated in three staged calls per problem,
L4 $\to$ L3 $\to$ L2, so that each later stage sees the already generated
more abstract levels and can respect the adjacent boundary.  Each call
receives the problem and the reference solution (truncated to 4,000
characters and marked as ``for your understanding only'').

The primary
Qwen3.5-397B-A17B compiler runs at temperature 0.3 with a 512-token output
cap.  The Qwen3-8B replication uses the same prompt texts and staging but
enables Qwen3 thinking mode at temperature 0.6, top-$p$ 0.95, and top-$k$ 20;
only final response content is retained, with an 8,192-token ceiling to leave
room for internal reasoning.  Both variants retry up to three times on
transport errors or empty output, but never on length.  Because model and
decoding mode change together, the replication tests a deployable compiler
package rather than isolating parameter count.

The prompts impose no hard
length constraint; they give a soft ``typical range'' and explicitly permit
deviation, so length varies with the abstraction level and
Table~\ref{tab:hint-lengths} reports the realized distributions.  Each
prompt also contains positive and negative worked examples and mandatory
self-check tests that the model must apply before answering.  We summarize
the binding constraints below; the complete prompt texts, including worked
examples, are in the code repository.

\begin{promptbox}[title={L4 (problem category; generated first)}]
Required pattern: ``<Domain>: <noun phrase describing the object / concept
type>.''; the noun phrase describes WHAT is asked, not how to solve it.\\
Must not contain: any solution-action verb (solve, compute, prove, derive,
convert, use, apply, calculate, count, find, determine, \ldots); any method,
technique, or theorem name; any thinking pattern or cognitive move; any
concrete numerical value or the final answer; any prepositional clause that
smuggles in a method (``via X'', ``using X'', ``through X'', ``by X'',
``based on X'', \ldots).\\
Self-checks: verb test; ``via X''-removal test (delete the clause; if the
remaining sentence no longer classifies the problem, the classification was
by method and must be rewritten); method-name test.\\
Typical length 10--30 tokens; brevity is a feature of L4, not a flaw.
\end{promptbox}

\begin{promptbox}[title={L3 (method-independent framing; sees L4)}]
Must be a single cognitive framing shift (e.g., shift to the complement,
unify into a common representation, seek an invariant), not an operation
sequence.\\
Must not contain: any named method, theorem, or algorithm; two or more
concrete action verbs (a verb-count test flags an operation sequence in
disguise); the words ``then'', ``next'', ``after'', ``subsequently''; any
concrete numerical value or the final answer.\\
Transferability self-check: the same sentence must apply unchanged to a
different problem in the same L4 category that requires a different concrete
method.\\
Typical length 20--60 tokens.
\end{promptbox}

\begin{promptbox}[title={L2 (named strategy; sees L4 and L3)}]
Must name at least one specific method, theorem, or tool that a student
could look up, and may add one short conceptual clause explaining why it
applies.\\
Must not contain: procedural language (``first X, then Y'', numbered steps,
``write down'', ``carry over''); any concrete numerical value from the
problem; any symbolic substitution (``let $x=\ldots$''); any intermediate
expression or the final answer.  ``Then'' is permitted only to connect two
named methods, never two procedural steps.\\
Self-checks: sequence-word ban; named-method requirement.\\
Typical length 30--90 tokens.
\end{promptbox}

The matched semantic audits of Appendix~\ref{app:audits} evaluate both
compilers' realized hints against these definitions on the same 1,200
problems.

\subsection{Teacher Interfaces}
\label{app:bridge-prompts}

Every teacher prompt in the primary runs instantiates one fixed template.
With \texttt{\{problem\}} and \texttt{\{hint\}} denoting the training fields,
the teacher-side user message is
\begin{promptbox}
Problem: \{problem\}

\medskip
Here is a \{label\} for this problem:\\
\{begin delimiter\}\\
\{hint\}\\
\{end delimiter\}\\
\{transition\}
\end{promptbox}
The student-side user message is identical for all conditions and never
contains the hint:
\begin{promptbox}
Problem: \{problem\}

\medskip
Please reason step by step, and put your final answer within
\textbackslash boxed\{\}.
\end{promptbox}
Table~\ref{tab:bridge-labels} lists the level-specific label and delimiters,
and the five transition texts follow.  The shared-bridge control replaces
all three elements with the single template of
Appendix~\ref{app:shared-bridge}, and the routed mixtures reuse the
level-specific elements below row by row.

\begin{table}[!htbp]
\caption{Level-specific intro label and hint delimiters.  The intro line is
always ``Here is a \{label\} for this problem:''.}
\label{tab:bridge-labels}
\centering
\small
\begin{tabular}{@{}lll@{}}
\toprule
Level & Label & Delimiter pair \\
\midrule
\LI & reference solution & \texttt{=== Reference Solution Begin/End ===} \\
\LII & strategy hint & \texttt{=== Strategy Hint Begin/End ===} \\
\LIII & thinking direction & \texttt{=== Thinking Direction Begin/End ===} \\
\LIV & problem category & \texttt{=== Problem Category Begin/End ===} \\
\LV & known final answer & \texttt{=== Final Answer Begin/End ===} \\
\bottomrule
\end{tabular}
\end{table}

All five transitions end with the same closing sentence, quoted once here and
elided as \textsc{[close]} below:
\begin{promptbox}[title={Shared closing, all levels}]
Think step by step, explore different approaches, and don't be afraid to
backtrack or reconsider if something doesn't work out:
\end{promptbox}
The L1--L4 openings are parallel and differ only in the level-appropriate verb
chain (\emph{understand--arrive}, \emph{understand--expand},
\emph{grasp--discover}, \emph{recall--select}); L5 instead directs the teacher
to construct a derivation toward a given destination.
\begin{promptbox}[title={L1 transition}]
After reading the reference solution above, make sure you truly understand
the reasoning behind each step --- do not copy or paraphrase it. Now, using
your own words and independent reasoning, arrive at the correct final
answer to the problem above. \textsc{[close]}
\end{promptbox}

\begin{promptbox}[title={L2 transition}]
After reading the strategy hint above, make sure you truly understand why
this strategy fits the problem --- do not copy or paraphrase it. Now, using
your own words and independent reasoning, expand this strategy into a full
derivation and arrive at the final answer to the problem above.
\textsc{[close]}
\end{promptbox}

\begin{promptbox}[title={L3 transition}]
After reading the thinking direction above, make sure you truly grasp how
this direction applies to the problem --- do not copy or paraphrase it. Now,
using your own words and independent reasoning, discover a concrete method
along this direction and arrive at the final answer to the problem above.
\textsc{[close]}
\end{promptbox}

\begin{promptbox}[title={L4 transition}]
After reading the problem category above, make sure you truly recall the
standard techniques used for this type of problem --- do not copy or
paraphrase it. Now, using your own words and independent reasoning, select a
suitable technique and derive the final answer to the problem above.
\textsc{[close]}
\end{promptbox}

\begin{promptbox}[title={L5 transition}]
The known final answer above gives only the destination, not the derivation.
Do not merely repeat or cite it. Now independently construct a complete,
valid reasoning path from the problem to that answer, justifying every
necessary step. \textsc{[close]}
\end{promptbox}

\subsection{Context Audits}
\label{app:audits}

Table~\ref{tab:hint-lengths} reports exact counts under the Qwen3 tokenizer,
which is shared across scales.  We report both stored-hint length and the
privileged increment, defined as the teacher prompt minus its matched no-hint
prompt, because the fixed bridge is a substantial fraction of short contexts.
All levels contain 29,434 aligned, nonempty rows with no error sentinels; L5
includes seven proof-target overrides.  These reductions concern the
teacher-scoring interface once a solution exists, not solution acquisition or
end-to-end wall-clock cost.

\begin{table*}[!htbp]
\caption{Exact Qwen3-token lengths over all 29,434 training examples.  P10
and P90 apply to raw hint content; the last two columns are means.  Privileged
increment includes the hint and its level-specific bridge but excludes the
problem and shared chat-template tokens.}
\label{tab:hint-lengths}
\centering
\small
\setlength{\tabcolsep}{4pt}
\begin{tabular*}{\textwidth}{@{\extracolsep{\fill}}lcccccc@{}}
\toprule
Context & Hint mean & Median & P10 & P90 & Teacher prompt & Privileged increment \\
\midrule
\LI: solution      & 657.3 & 630 & 322 & 1,022 & 835.0 & 734.4 \\
\LII: strategy      & 38.9  & 38  & 31  & 48    & 225.5 & 124.9 \\
\LIII: framing       & 25.2  & 25  & 20  & 32    & 212.7 & 112.2 \\
\LIV: category      & 17.7  & 18  & 14  & 22    & 203.3 & 102.7 \\
\LV: answer/target & 6.7   & 4   & 1   & 15    & 183.9 & 83.4 \\
\bottomrule
\end{tabular*}
\end{table*}

\begin{figure*}[!htbp]
\centering
\includegraphics[width=0.9\textwidth]{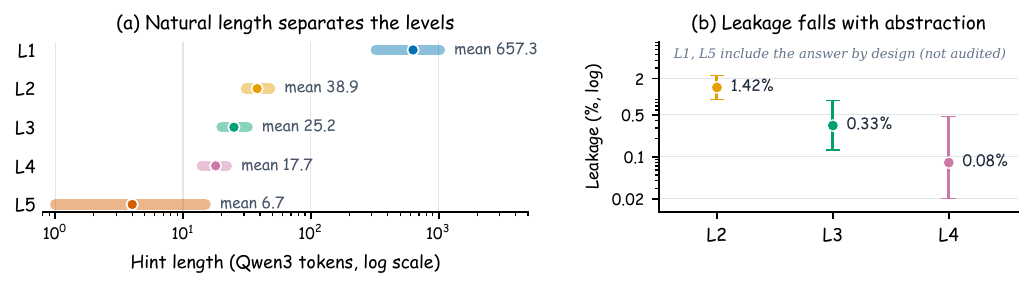}
\caption{Primary-compiler context length and answer-leakage audit.
\textbf{(a)} Hint length across all 29,434 problems, measured in Qwen3 tokens
and shown on a log scale; Table~\ref{tab:hint-lengths} reports summary
statistics.
\textbf{(b)} Adjudicated answer-equivalent leakage on a stratified sample of
1,200 problems.  L1 and L5 contain the answer by design;
Table~\ref{tab:hint-audit} reports matched audits of both compilers.}
\label{fig:intervention-audit}
\end{figure*}

The deterministic scan covers all 29,434 rows from each compiler.  For
semantic properties, we draw 1,200 unique problem rows with seed 20260719,
proportionally stratified
over four answer types (numeric, expression or text, multiple choice,
and proof target) and four reference-solution length quartiles.  The same rows
are used for both compilers, each contributing 3,600 audited hints
($1{,}200$ problems $\times$ three levels).
The audit distinguishes \emph{direct leakage}, which explicitly states the
final value, expression, option, or answer, from \emph{answer-equivalent
leakage}, which states a conclusion that determines the final answer without
carrying out the intended solution.  A useful method alone is not counted as
leakage.  Boundary compliance additionally requires L2 to name a concrete
method without executing a derivation, L3 to provide one method-independent
framing without an operation sequence, and L4 to identify the correct domain
and objects without recommending an action.

Two zero-temperature prompt variants independently annotate every
compiler--problem triplet.  Of the triplets with at least one disagreement, we
send the 31 from the primary compiler and the 67 from the Qwen3-8B compiler to
a third adjudication pass using the same inputs without access to the earlier
judgments.  This resolves all 2,400
compiler--problem records, with no unresolved cases.  All three passes use Qwen3.5-397B-A17B,
so the resulting agreement is prompt-replication agreement from one model
rather than human inter-rater agreement.

\begin{table*}[!htbp]
\caption{Matched semantic audits of both compilers on the same 1,200 problem
rows after adjudication.  Rates are computed over the 1,200 hints at each
compiler--level pair.  ``Core role valid'' checks concrete-method presence for
L2, single-framing presence for L3, and both domain and object correctness for
L4.}
\label{tab:hint-audit}
\centering
\small
\setlength{\tabcolsep}{4pt}
\begin{tabular*}{\textwidth}{@{\extracolsep{\fill}}llcccc@{}}
\toprule
Compiler & Level & Direct & Answer-equivalent & Boundary pass & Core role valid \\
\midrule
\multirow{3}{*}{397B primary}
 & \LII: strategy & 0/1200 & 17/1200 (1.42\%) & 98.58\% & 100.00\% \\
 & \LIII: framing  & 0/1200 &  4/1200 (0.33\%) & 99.67\% & 100.00\% \\
 & \LIV: category & 0/1200 &  1/1200 (0.08\%) & 99.92\% & 100.00\% \\
\midrule
\multirow{3}{*}{8B replication}
 & \LII: strategy & 0/1200 & 50/1200 (4.17\%) & 95.58\% & 99.92\% \\
 & \LIII: framing  & 0/1200 & 31/1200 (2.58\%) & 97.08\% & 99.83\% \\
 & \LIV: category & 0/1200 &  2/1200 (0.17\%) & 98.83\% & 99.17\% \\
\bottomrule
\end{tabular*}
\end{table*}

No hint directly copies an answer.  The 8B compiler has higher L2/L3 leakage
and boundary-failure rates, whereas its L4 context remains low-leakage and
follows the scale-dependent direction in Table~\ref{tab:compiler-scale}.  The
L4 result shows that the finding is not confined to the leakier L2/L3
contexts; leakage may still affect individual L2/L3 cells.

Table~\ref{tab:hint-audit-agreement} reports agreement for the two fields with
positive adjudicated cases.
\begin{table*}[!htbp]
\caption{Prompt-A/B agreement on the two audit fields that contain positive
adjudicated cases.  For primary-compiler L4, $\kappa=0$ is a prevalence
artifact: one prompt assigns no positives and the other assigns only two,
despite 99.83\% raw agreement.}
\label{tab:hint-audit-agreement}
\centering
\small
\setlength{\tabcolsep}{4pt}
\begin{tabular*}{\textwidth}{@{\extracolsep{\fill}}llrrrr@{}}
\toprule
& & \multicolumn{2}{c}{397B primary} & \multicolumn{2}{c}{8B replication} \\
\cmidrule(lr){3-4}\cmidrule(lr){5-6}
Level & Field & Agreement & $\kappa$ & Agreement & $\kappa$ \\
\midrule
\LII & Answer leakage & 97.83\% & 0.527 & 96.75\% & 0.695 \\
\LII & Boundary pass  & 97.67\% & 0.507 & 96.33\% & 0.680 \\
\LIII & Answer leakage & 99.58\% & 0.614 & 98.17\% & 0.694 \\
\LIII & Boundary pass  & 99.58\% & 0.614 & 97.92\% & 0.673 \\
\LIV & Answer leakage & 99.83\% & 0.000 & 100.00\% & 1.000 \\
\LIV & Boundary pass  & 99.83\% & 0.000 & 99.33\% & 0.711 \\
\bottomrule
\end{tabular*}
\end{table*}

The audit release accompanying the code will include sample IDs, prompts, raw
outputs, adjudications, and aggregation code.

The two replicated audit passes and the adjudication pass share one block of
operational definitions and differ only in the role preamble:
\begin{promptbox}
(A) You are a conservative data-quality auditor for mathematical reasoning
supervision. Apply the operational definitions literally and do not infer
unstated content.

\medskip
(B) You are an adversarial compliance reviewer. Look independently for subtle
answer-equivalent leakage and boundary drift, but do not penalize a hint
merely because it is useful. Evaluate each level against its absolute
definition, not relative to the other hints.

\medskip
(C) You are the final independent adjudicator for a mathematical hint audit.
Resolve the case from the source text alone; you are not shown earlier
judgments. Be strict about evidence and use `uncertain' only when the source
wording truly admits two readings.
\end{promptbox}
The shared operational definitions are:
\begin{promptbox}
Audit the three hints against their declared definitions. Judge only what is
actually written in each hint, using the problem, reference solution, and
ground-truth answer only as verification context.

\medskip
ANSWER LEAKAGE:\\
- ``direct'': explicitly states the final value, expression, option, or
answer.\\
- ``equivalent'': states a conclusion algebraically/logically equivalent to
the final answer, so the answer is already determined without carrying out
the intended solution.\\
- ``none'': contains only guidance or classification. A useful strategy that
makes the problem easier is NOT leakage.\\
- ``uncertain'': evidence is genuinely ambiguous.

\medskip
L2 (named strategy): Must identify a concrete method, theorem, rule,
representation, or tool. May briefly say why it fits. Must not carry out
problem-specific calculations or a derivation. Must not give a multi-step
operation sequence or the answer.

\medskip
L3 (method-independent framing): Must express one transferable cognitive
framing shift. Must not name a specific theorem, algorithm, or concrete
method. Must not prescribe a sequence of operations or give the answer.

\medskip
L4 (problem category): Must correctly identify the mathematical domain and
relevant object/concept. Must not recommend a method, cognitive move, action
sequence, or answer.

\medskip
For every false/violation judgment, quote at most two short exact spans from
the corresponding hint in ``evidence''. Do not quote the reference solution.
Set boundary\_pass from the complete definition, not merely from answer
leakage. Return JSON only, with exactly the requested schema.
\end{promptbox}
The user message supplies the sample ID, problem, ground-truth answer, the
reference solution (as verification context, clipped to 24,000 characters),
the paired L2/L3/L4 hints, and the required JSON schema.  Each level is
annotated with \texttt{answer\_leakage}
$\in\{$\texttt{none}, \texttt{direct}, \texttt{equivalent},
\texttt{uncertain}$\}$, level-specific boolean boundary fields (L2:
\texttt{concrete\_method\_present}, \texttt{problem\_specific\_execution},
\texttt{multi\_step\_sequence}; L3: \texttt{single\_framing\_present},
\texttt{named\_method\_present}, \texttt{operation\_sequence}; L4:
\texttt{domain\_correct}, \texttt{objects\_correct},
\texttt{method\_or\_action\_present}), an overall \texttt{boundary\_pass}, a
\texttt{confidence} grade, and at most two verbatim evidence spans of at most
240 characters.  Responses that violate this schema are rejected and retried,
so every retained annotation is structurally valid.

\section{Full Results and Robustness}

\subsection{Full Evaluation Results}

Table~\ref{tab:ood-main} gives the complete transfer results at step 200
summarized in Section~\ref{sec:results}, and Table~\ref{tab:zebra-details}
expands ZebraLogic into cell accuracy and the Hard and XL subsets.  Both are
single-run comparisons at a common endpoint; the multi-seed analysis covers
the L1, L3, and L4 in-domain comparison (Appendix~\ref{app:multi-seed}).
Table~\ref{tab:ceval-32k} reports the separately interpreted C-Eval audit,
discussed at the end of this subsection.

\begin{table*}[!htbp]
\caption{Transfer using the fixed checkpoint at step 200, with the primary compiler for L2--L4. Scores are percentages: MT-AIME reports macro Avg@12 over seven languages, GPQA Avg@10, AutoLogi verifier accuracy, and ZebraLogic exact puzzle accuracy. L5 is the answer-only control. Shading marks the highest and second-highest point estimates within each scale and benchmark, including ties.}
\label{tab:ood-main}
\centering
\small
\setlength{\tabcolsep}{4pt}
\begin{tabular*}{\textwidth}{@{\extracolsep{\fill}}llcccc@{}}
\toprule
Model & Condition & MT-AIME-7Lang & AutoLogi & GPQA-Diamond & ZebraLogic \\
\midrule
\multirow{6}{*}{1.7B}
 & Base & 42.62 & 86.73 & 36.72 & 61.20 \\
 & \LI & 46.94 & \second{87.43} & \second{38.54} & \best{65.10} \\
 & \LII & 46.79 & 86.22 & 37.98 & 62.60 \\
 & \LIII & \best{48.65} & 86.54 & \best{39.34} & \second{64.40} \\
 & \LIV & \second{47.86} & \best{87.62} & 37.73 & 62.70 \\
 & \LV & 44.84 & 86.54 & 38.18 & 61.30 \\
\midrule
\multirow{6}{*}{4B}
 & Base & 65.87 & 91.75 & 53.69 & 82.20 \\
 & \LI & 67.46 & 92.38 & 54.75 & 82.50 \\
 & \LII & \second{68.85} & 92.13 & \best{55.96} & 82.10 \\
 & \LIII & 67.62 & \second{92.63} & 55.25 & \second{82.60} \\
 & \LIV & \best{69.56} & 92.25 & \second{55.51} & \best{82.90} \\
 & \LV & 68.21 & \best{92.76} & 53.28 & 82.40 \\
\midrule
\multirow{6}{*}{8B}
 & Base & 70.48 & \second{92.13} & 60.05 & 85.60 \\
 & \LI & 71.90 & \best{92.63} & 61.26 & 86.40 \\
 & \LII & \best{73.61} & \second{92.13} & 60.45 & 87.30 \\
 & \LIII & 72.70 & 91.87 & \best{62.68} & \best{88.10} \\
 & \LIV & \second{72.74} & \second{92.13} & 60.51 & \second{87.60} \\
 & \LV & 71.43 & 91.94 & \second{61.67} & 85.90 \\
\bottomrule
\end{tabular*}
\end{table*}

\begin{table*}[!htbp]
\caption{Complete ZebraLogic-grid results at step 200.  Puzzle is exact-grid
accuracy; Cell measures individual grid entries.  Hard and XL are
exact-puzzle accuracy on the corresponding subsets.  $\Delta$ from Base is
the absolute percentage-point change from the same-scale unadapted Base
checkpoint.  Light-blue/bold shading marks the best and
light-gray/underline shading the second-best condition within each model
scale.}
\label{tab:zebra-details}
\centering
\small
\setlength{\tabcolsep}{4pt}
\begin{tabular*}{\textwidth}{@{\extracolsep{\fill}}llccccc@{}}
\toprule
Model & Condition & Puzzle & $\Delta$ from Base & Cell & Hard & XL \\
\midrule
\multirow{6}{*}{1.7B}
 & Base & 61.20 & -- & 68.88 & 48.06 & 7.00 \\
 & \LI & \best{65.10} & \best{+3.90} & \second{72.74} & \best{53.47} & \best{10.50} \\
 & \LII & 62.60 & +1.40 & 70.41 & 49.72 & 8.00 \\
 & \LIII & \second{64.40} & \second{+3.20} & \best{73.09} & \second{53.06} & 7.50 \\
 & \LIV & 62.70 & +1.50 & 71.48 & 50.42 & \second{9.00} \\
 & \LV & 61.30 & +0.10 & 69.84 & 47.92 & 6.00 \\
\midrule
\multirow{6}{*}{4B}
 & Base & 82.20 & -- & 84.21 & 75.42 & \second{36.50} \\
 & \LI & 82.50 & +0.30 & 84.83 & 75.97 & 36.00 \\
 & \LII & 82.10 & $-0.10$ & 83.14 & 75.42 & 34.00 \\
 & \LIII & \second{82.60} & \second{+0.40} & \second{85.38} & \second{76.25} & 36.00 \\
 & \LIV & \best{82.90} & \best{+0.70} & \best{85.91} & \best{76.53} & 36.00 \\
 & \LV & 82.40 & +0.20 & 84.78 & 75.69 & \best{37.50} \\
\midrule
\multirow{6}{*}{8B}
 & Base & 85.60 & -- & 86.80 & 80.14 & 41.00 \\
 & \LI & 86.40 & +0.80 & 86.85 & 81.53 & 44.50 \\
 & \LII & 87.30 & +1.70 & 87.45 & 82.64 & 46.00 \\
 & \LIII & \best{88.10} & \best{+2.50} & \best{88.25} & \best{83.89} & \second{46.50} \\
 & \LIV & \second{87.60} & \second{+2.00} & \second{87.83} & \second{83.06} & \best{47.50} \\
 & \LV & 85.90 & +0.30 & 86.13 & 80.97 & 43.50 \\
\bottomrule
\end{tabular*}
\end{table*}

C-Eval is a bounded-generation audit rather than a reasoning-transfer
benchmark, so it is read on its own.  Under a 32,768-token output budget, no
training condition improves on the unadapted Base consistently across scales on
C-Eval Macro (Table~\ref{tab:ceval-32k}).  Base is the best condition at 1.7B (69.82 against
69.23 for L1) and at 8B (84.60 against 84.57 for L5); the one condition that
exceeds Base anywhere is the answer-only control L5 at 4B, by 0.07 points
(80.80 against 80.73), while none of L1--L4 exceeds Base at any scale.
Extraction failures rise under every training condition, most sharply at 1.7B,
from 0.68\% for Base to 3.70\% for L2, and the metric charges each failure as
an error.  Under this protocol, therefore, OPSD training on competition
mathematics yields no consistent gain on broad Chinese knowledge questions, and
it leaves more responses without a parseable answer.  This is why C-Eval
is kept apart from the four reasoning-transfer benchmarks in the main text.

\begin{table*}[!htbp]
\caption{C-Eval test results with a 32,768-token output budget.  Macro is
the mean over 52 subjects; Hard Macro covers the official C-Eval Hard
subjects.  Extraction failures are counted as incorrect in Macro and Micro.
Light-blue/bold shading marks the best and light-gray/underline shading the
second-best condition within each model scale.}
\label{tab:ceval-32k}
\centering
\small
\setlength{\tabcolsep}{4pt}
\begin{tabular*}{\textwidth}{@{\extracolsep{\fill}}llcccc@{}}
\toprule
Model & Condition & Macro & Micro & Hard Macro & Extraction fail. \\
\midrule
\multirow{6}{*}{1.7B}
 & Base & \best{69.82} & \best{67.97} & \best{73.77} & \best{0.68} \\
 & \LI & \second{69.23} & \second{67.49} & 73.19 & \second{1.67} \\
 & \LII & 67.46 & 65.78 & 72.31 & 3.70 \\
 & \LIII & 67.86 & 66.25 & 73.35 & 3.04 \\
 & \LIV & 67.78 & 66.11 & \second{73.75} & 3.48 \\
 & \LV & 68.49 & 66.85 & 72.68 & 2.43 \\
\midrule
\multirow{6}{*}{4B}
 & Base & \second{80.73} & \best{79.40} & \second{85.91} & \best{0.77} \\
 & \LI & 80.03 & 78.73 & 84.96 & 2.26 \\
 & \LII & 80.11 & 78.78 & 84.45 & 2.31 \\
 & \LIII & 80.09 & 78.76 & 85.21 & 2.21 \\
 & \LIV & 80.52 & 79.14 & 85.18 & \second{1.77} \\
 & \LV & \best{80.80} & \second{79.37} & \best{86.28} & 1.96 \\
\midrule
\multirow{6}{*}{8B}
 & Base & \best{84.60} & \best{83.61} & 88.45 & \best{0.84} \\
 & \LI & 84.43 & 83.37 & \second{88.47} & 1.27 \\
 & \LII & 84.13 & 82.98 & 87.75 & 1.43 \\
 & \LIII & 84.31 & 83.26 & 88.09 & 1.51 \\
 & \LIV & 84.53 & \second{83.46} & 87.89 & 1.26 \\
 & \LV & \second{84.57} & 83.41 & \best{89.40} & \second{1.17} \\
\bottomrule
\end{tabular*}
\end{table*}

\subsection{Context Controls}
\label{app:compiler-replication}
\label{app:prefix-control}

Two controls vary how the privileged context is produced while holding the
training pipeline fixed.  Table~\ref{tab:compiler-scale} regenerates L2--L4 with
Qwen3-8B under the same semantic contracts, so the ordering between full and
intermediate contexts can be checked against a second compiler.
Table~\ref{tab:prefix-ablation} replaces semantic rewriting with a fixed
10\% reference prefix plus the answer at 4B, which shortens the context
without abstracting it; that comparison matches L2--L4 in neither length nor
answer access.  Section~\ref{sec:controls} discusses both results.

\begin{table}[t]
\caption{Compiler replication using ID peak means. L2--L4 are generated with the primary Qwen3.5-397B-A17B compiler or with Qwen3-8B; L1 requires no compiler and is the shared reference at each scale for both halves, not an additional run. Bold marks the best condition within each compiler block, counting the shared L1 column. The peak ordering between full and intermediate contexts is preserved across compilers, while the preferred intermediate level can change.}
\label{tab:compiler-scale}
\centering
\small
\begingroup
\renewcommand{\arraystretch}{1.08}
\setlength{\tabcolsep}{4pt}
\begin{tabular}{@{}lccccccc@{}}
\toprule
\multirow{2}{*}{Student} & \multirow{2}{*}{\LI} & \multicolumn{3}{c}{397B primary compiler} & \multicolumn{3}{c}{Qwen3-8B compiler} \\
\cmidrule(lr){3-5} \cmidrule(lr){6-8}
& & \LII & \LIII & \LIV & \LII & \LIII & \LIV \\
\midrule
1.7B & \textbf{43.89} & 40.83 & 42.78 & 43.61 & 41.02 & 42.78 & 42.96 \\
4B & 63.43 & 64.07 & \textbf{64.81} & 64.63 & 64.07 & 63.71 & \textbf{64.35} \\
8B & 66.67 & \textbf{68.24} & 66.85 & 67.22 & \textbf{67.96} & 67.04 & 67.13 \\
\bottomrule
\end{tabular}
\endgroup
\end{table}

\begin{table}[t]
\caption{Comparison of semantic contexts and prefix truncation with Qwen3-4B. ID peak mean averages each benchmark's best Avg@12 across eight checkpoints, while ID@200 uses step 200. Only the prefix condition is new; the L1--L5 rows repeat the 4B primary runs, whose per-benchmark peaks are in Table~\ref{tab:id-main}. The upper block includes the final answer, the lower block prohibits it by design. The comparison matches neither length nor answer access.}
\label{tab:prefix-ablation}
\centering
\small
\setlength{\tabcolsep}{4pt}
\begin{tabular*}{\columnwidth}{@{\extracolsep{\fill}}lcc@{}}
\toprule
Condition & ID peak mean & ID@200 \\
\midrule
\LI: complete solution & 63.43 & 60.93 \\
\LV: answer only & 63.61 & \second{62.50} \\
Prefix-10\% + answer & 63.89 & 61.67 \\
\midrule
\LII: strategy & 64.07 & \best{62.59} \\
\LIII: framing & \best{64.81} & 61.76 \\
\LIV: category & \second{64.63} & \best{62.59} \\
\bottomrule
\end{tabular*}
\end{table}

\subsection{Training Robustness}
\label{app:budget-robustness}

Three checks probe whether the granularity contrast depends on the training
budget, the optimization seed, or the checkpoint at which runs are read.

The primary experiments fix the training rollout budget at 1,024 tokens,
following the original \opsd{} configuration.  Because sampled completions
frequently reach this cap, we verify that the granularity comparison
survives a doubled budget: L1--L4 are retrained at 4B with a 2,048-token
rollout budget and otherwise identical configuration, then evaluated under
the standard in-domain protocol.

\begin{table}[!htbp]
\caption{Robustness to rollout budget at 4B (Avg@12 means across three benchmarks).
Peak takes each benchmark's best checkpoint; single-ckpt selects the one
checkpoint with the best mean across three benchmarks (step in parentheses); @200
is the fixed endpoint.  Truncation reports the fraction of training
rollouts reaching the token cap early (steps $\le$50) and late
(steps $>$150) in training.  Bold marks the column maximum, including ties.
Intermediate levels match or beat L1 on the
peak and single-checkpoint means under both budgets.}
\label{tab:budget-robustness}
\centering
\small
\setlength{\tabcolsep}{3.5pt}
\begin{tabular}{@{}lrrrrrrrr@{}}
\toprule
& \multicolumn{4}{c}{1,024-token rollouts} & \multicolumn{4}{c}{2,048-token rollouts} \\
\cmidrule(lr){2-5} \cmidrule(l){6-9}
Level & Peak & Single-ckpt & @200 & Trunc.\ e/l & Peak & Single-ckpt & @200 & Trunc.\ e/l \\
\midrule
\LI & 63.43 & 62.31 (125) & 60.93 & 57\%/77\% & 63.43 & 63.06 (50) & 60.93 & 21\%/58\% \\
\LII & 64.07 & 63.43 (125) & \best{62.59} & 56\%/83\% & 64.54 & 64.26 (75) & 58.61 & 21\%/67\% \\
\LIII & \best{64.81} & 63.24 (150) & 61.76 & 57\%/81\% & 64.35 & 63.70 (175) & 61.48 & 22\%/52\% \\
\LIV & 64.63 & \best{64.63} (125) & \best{62.59} & 55\%/77\% & \best{65.37} & \best{64.63} (75) & \best{61.57} & 22\%/54\% \\
\bottomrule
\end{tabular}
\end{table}

Table~\ref{tab:budget-robustness} shows that the doubled budget roughly halves
truncation early in training, yet changes each level's peak mean by at most 0.74
points.  Every intermediate context again matches or exceeds L1, while the
nominal leader shifts from L3 to L4.  The larger budget moves three of four
single-checkpoint peaks earlier and does not improve any endpoint at step 200,
suggesting a change in effective training horizon while preserving the
family-level ordering.  The endpoint is more sensitive to the budget: L2's
mean at step 200 falls from 62.59 to 58.61 even though its peak mean is higher
under the larger budget (64.07 to 64.54), so preserving the peak does not
imply preserving the endpoint.  These runs use one seed at 4B and do not
evaluate transfer.

\phantomsection
\label{app:multi-seed}
To probe optimization variance directly, L1, L3, and L4 are retrained at all
three scales with two additional configurations under
Table~\ref{tab:training-config}.  The primary run uses the default seed 42;
the additional runs use seeds 43 and 44, with the trainer, data order, and
rollout RNGs changed together and disjoint across configurations.  Every run is evaluated with the standard
in-domain protocol.  Table~\ref{tab:multi-seed} groups the per-benchmark
bests, the peak mean across three benchmarks, and the endpoint at step 200 by
configuration, so each block compares runs with the same seed configuration, and reports
the paired peak difference from that configuration's own L1 run.

\begin{table}[!htbp]
\caption{Replication across training seeds at all three scales (Avg@12).  ID peak
takes the best checkpoint per benchmark up to step 200 and averages the three;
ID@200 is the selection-free endpoint mean.  Rows are grouped by seed
configuration, so every block compares intermediate contexts
against its own L1 run; seed 42 is the primary run of
Table~\ref{tab:id-main}, and 43 and 44 are the two additional configurations.
$\Delta$ is the ID peak difference from the L1 run in the same block.
The best scores are not shaded because the configurations do not agree on an
intermediate leader.}
\label{tab:multi-seed}
\centering
\small
\setlength{\tabcolsep}{4pt}
\begin{tabular}{@{}lllrrrrrr@{}}
\toprule
Student & Seed & Level & AIME24 & AIME25 & HMMT25 & ID peak & $\Delta$ vs.\ \LI & ID@200 \\
\midrule
\multirow{9}{*}{1.7B}
 & \multirow{3}{*}{42} & \LI & 57.22 & 43.33 & 31.11 & 43.89 & --- & 42.31 \\
 & & \LIII & 56.94 & 41.39 & 30.00 & 42.78 & $-1.11$ & 40.00 \\
 & & \LIV & 59.72 & 43.06 & 28.06 & 43.61 & $-0.28$ & 39.26 \\
\cmidrule(lr){2-9}
 & \multirow{3}{*}{43} & \LI & 55.28 & 43.89 & 28.61 & 42.59 & --- & 38.89 \\
 & & \LIII & 57.22 & 41.67 & 30.00 & 42.96 & $+0.37$ & 40.74 \\
 & & \LIV & 58.06 & 42.22 & 28.61 & 42.96 & $+0.37$ & 39.81 \\
\cmidrule(lr){2-9}
 & \multirow{3}{*}{44} & \LI & 56.39 & 41.39 & 29.72 & 42.50 & --- & 41.11 \\
 & & \LIII & 58.33 & 41.39 & 28.89 & 42.87 & $+0.37$ & 39.07 \\
 & & \LIV & 57.50 & 41.94 & 27.78 & 42.41 & $-0.09$ & 39.26 \\
\midrule
\multirow{9}{*}{4B}
 & \multirow{3}{*}{42} & \LI & 76.94 & 68.61 & 44.72 & 63.43 & --- & 60.93 \\
 & & \LIII & 77.50 & 69.72 & 47.22 & 64.81 & $+1.39$ & 61.76 \\
 & & \LIV & 76.94 & 69.17 & 47.78 & 64.63 & $+1.20$ & 62.59 \\
\cmidrule(lr){2-9}
 & \multirow{3}{*}{43} & \LI & 76.67 & 69.44 & 45.28 & 63.80 & --- & 60.83 \\
 & & \LIII & 76.39 & 68.89 & 46.11 & 63.80 & $0.00$ & 63.06 \\
 & & \LIV & 76.94 & 70.28 & 46.67 & 64.63 & $+0.83$ & 62.78 \\
\cmidrule(lr){2-9}
 & \multirow{3}{*}{44} & \LI & 76.11 & 68.06 & 46.11 & 63.43 & --- & 61.39 \\
 & & \LIII & 76.67 & 70.00 & 45.83 & 64.17 & $+0.74$ & 63.33 \\
 & & \LIV & 76.94 & 69.72 & 46.11 & 64.26 & $+0.83$ & 60.93 \\
\midrule
\multirow{9}{*}{8B}
 & \multirow{3}{*}{42} & \LI & 78.89 & 73.06 & 48.06 & 66.67 & --- & 65.28 \\
 & & \LIII & 79.44 & 71.39 & 49.72 & 66.85 & $+0.19$ & 65.37 \\
 & & \LIV & 80.56 & 71.39 & 49.72 & 67.22 & $+0.56$ & 64.63 \\
\cmidrule(lr){2-9}
 & \multirow{3}{*}{43} & \LI & 78.61 & 73.06 & 47.78 & 66.48 & --- & 63.61 \\
 & & \LIII & 78.89 & 73.61 & 49.72 & 67.41 & $+0.93$ & 63.70 \\
 & & \LIV & 80.56 & 73.06 & 48.33 & 67.31 & $+0.83$ & 65.00 \\
\cmidrule(lr){2-9}
 & \multirow{3}{*}{44} & \LI & 79.17 & 73.06 & 48.33 & 66.85 & --- & 65.28 \\
 & & \LIII & 79.17 & 72.22 & 48.06 & 66.48 & $-0.37$ & 65.93 \\
 & & \LIV & 79.72 & 71.94 & 49.17 & 66.94 & $+0.09$ & 65.19 \\
\bottomrule
\end{tabular}
\end{table}

At 4B and 8B, L4 exceeds its matched L1 peak in all three configurations and
L3 does so in four of the six scale--configuration cells, falling level at 4B
seed~43 and below at 8B seed~44 (Table~\ref{tab:multi-seed}).  Aggregating
these runs under the four checkpoint rules of
Table~\ref{tab:aggregation-summary} preserves the direction at both scales,
including when all checkpoints are averaged.  At 1.7B the peak differences range from
$-1.11$ to $+0.37$ points and the selection-free rules favor L1 on average,
by 0.14 to 1.33 points.

The endpoint at step 200 is the most sensitive to seed among
the four rules, carrying the largest paired standard deviation in five of the
six contrasts; the exception is 8B L3, whose peak spread of 0.65 exceeds its
endpoint spread of 0.32.  Among L3 and
L4, the leader is not stable within any scale: at 4B, L3 leads the primary run
and L4 leads seeds 43 and 44; at 8B, L4 leads the primary run and seed 44 while
L3 leads seed 43.  These replications support the
contrast between L1 and intermediate contexts at the larger scales, while the
ordering among intermediate levels remains unstable.  They cover L1, L3, and L4; L2, L5,
the transfer evaluations, and the remaining controls are single runs.

\phantomsection
\label{app:training-curves}

\begin{figure*}[!htbp]
\centering
\includegraphics[width=\textwidth]{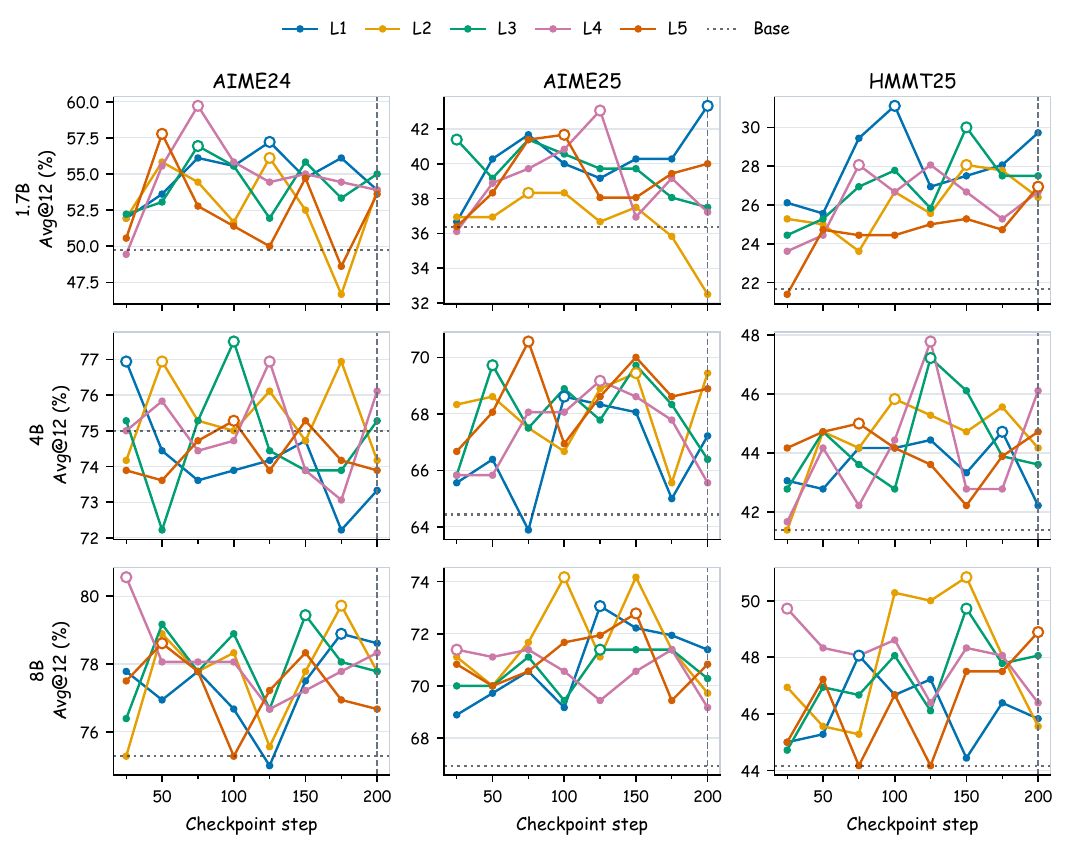}
\caption{In-domain Avg@12 vs.\ checkpoint step for each condition and
benchmark, primary run (seed 42), with the unadapted Base score shown for
reference.  Open circles mark each curve's maximum
through step 200; the
dashed vertical line marks the common endpoint used for transfer evaluation.
The substantial variation across checkpoints motivates separating
in-domain peak reporting from transfer at a fixed checkpoint.}
\label{fig:eval-vs-step}
\end{figure*}

Figure~\ref{fig:eval-vs-step} shows in-domain Avg@12 at every checkpoint
(steps 25, 50, \dots, 200) on AIME24, AIME25, and HMMT25, with open circles at
each curve's best checkpoint.  The benchmark curves are visibly nonmonotonic,
which is what motivates reporting peaks and selection-free endpoints side by
side.  Table~\ref{tab:id-steps} lists the step at which each peak in
Table~\ref{tab:id-main} occurs.  The selected steps are spread across the
schedule rather than concentrated at its end, which is the same observation
from the tabular side.

\begin{figure*}[!htbp]
\centering
\includegraphics[width=\textwidth]{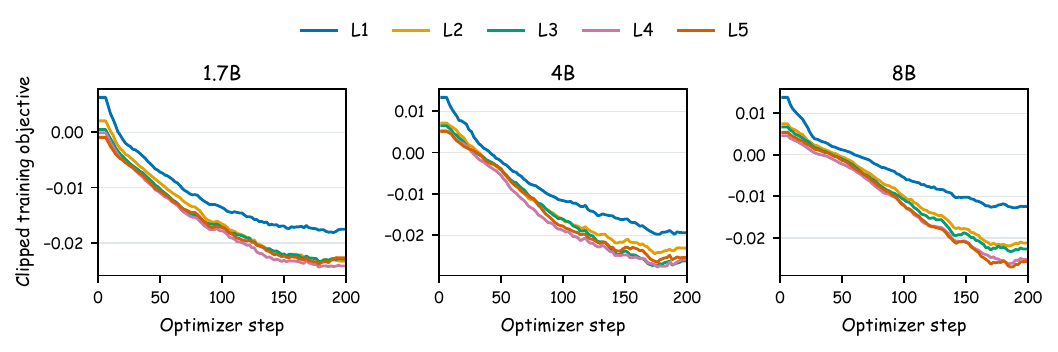}
\caption{Logged clipped training objective for each teacher context
condition and model scale, under a moving average over five records.  Each condition
has a different teacher target, so the vertical ordering describes the
respective objectives only and is comparable neither across conditions nor to
the downstream benchmark ranking.}
\label{fig:training-loss}
\end{figure*}

\begin{table}[!htbp]
\caption{Step at which each peak Avg@12 in Table~\ref{tab:id-main} is attained,
primary run (seed 42).  Checkpoints are taken every 25 steps up to 200.}
\label{tab:id-steps}
\centering
\small
\setlength{\tabcolsep}{4pt}
\begin{tabular*}{\columnwidth}{@{\extracolsep{\fill}}llccccc@{}}
\toprule
Model & Benchmark & \LI & \LII & \LIII & \LIV & \LV \\
\midrule
\multirow{3}{*}{1.7B}
 & AIME24 & 125 & 125 & 75 & 75 & 50 \\
 & AIME25 & 200 & 75 & 25 & 125 & 100 \\
 & HMMT25 & 100 & 150 & 150 & 75 & 200 \\
\midrule
\multirow{3}{*}{4B}
 & AIME24 & 25 & 50 & 100 & 125 & 100 \\
 & AIME25 & 100 & 150 & 50 & 125 & 75 \\
 & HMMT25 & 175 & 100 & 125 & 125 & 75 \\
\midrule
\multirow{3}{*}{8B}
 & AIME24 & 175 & 175 & 150 & 25 & 50 \\
 & AIME25 & 125 & 100 & 125 & 25 & 150 \\
 & HMMT25 & 75 & 150 & 150 & 25 & 200 \\
\bottomrule
\end{tabular*}
\end{table}  Figure~\ref{fig:training-loss} shows the logged clipped
teacher--student objective as an auxiliary trace.  Each condition optimizes
against a different teacher target on different on-policy rollouts, so its
values are not on a common scale across conditions and cannot be read as
evidence that the conditions trained equally well; the curves only indicate
that no run diverged.

For each (model, condition, benchmark) triple, define
$\Delta_{200} = \text{Avg@12}_{\text{best}\le 200} - \text{Avg@12}_{200}$
in percentage points.  Table~\ref{tab:checkpoint-gap} averages this quantity
over the three in-domain benchmarks.  Across all 45 triples, the median and
mean gaps are 1.94 and 2.31 points, respectively.  The largest is 5.83 points
for 1.7B-L4 on AIME25 (43.06 at step 125 versus 37.22 at step 200).  Thus,
step 200 is a common, selection-free endpoint and need not coincide with
convergence.  The OOD comparison remains internally matched because every
condition uses the same stopping rule; sensitivity to alternative fixed
endpoints remains open.

\begin{table}[!htbp]
\caption{Mean Avg@12 gap between the best checkpoint through step 200 and the endpoint at step 200
($\Delta_{200}$, percentage points), averaged over AIME24, AIME25, and
HMMT25.  Smaller is closer to the condition's best checkpoint.}
\label{tab:checkpoint-gap}
\centering
\small
\begin{tabular}{@{}lrrrrr@{}}
\toprule
Model & \LI & \LII & \LIII & \LIV & \LV \\
\midrule
1.7B & 1.57 & 3.24 & 2.78 & 4.35 & 1.95 \\
4B   & 2.50 & 1.48 & 3.06 & 2.04 & 1.11 \\
8B   & 1.39 & 3.89 & 1.48 & 2.59 & 1.30 \\
\bottomrule
\end{tabular}
\end{table}

\subsection{Scope of the Controls}
\label{app:additional-limitations}

This subsection records which controls bound the scope conditions summarized in
Section~\ref{sec:interpretation}.  Against single-run evidence: replication
across three seeds covers L1, L3, and L4 at all three scales
(Appendix~\ref{app:multi-seed}), while the prefix, bridge, and rollout budget
controls center on 4B and the mixture and routing studies are single runs.  Against selection by score: the
in-domain protocol picks the best of eight checkpoints separately per benchmark,
so we report selection-free endpoints and full trajectories in
Appendix~\ref{app:training-curves}.

Against bridge wording, which varies with
level in the primary runs: the shared-bridge control holds it fixed
(Appendix~\ref{app:shared-bridge}).  Against the confound between semantics and
length, which the levels vary together: the prefix control fixes truncation with
answer access, though it matches L2--L4 in neither length nor answer access
(Appendix~\ref{app:prefix-control}).  Against compiler identity: the Qwen3-8B
replication stays inside the Qwen family and shows higher L2/L3 leakage, and
both compilers receive the reference solution, so it broadens the compiler
evidence without isolating parameter count.

Against difficulty-bucket
artifacts: grouping by Base rather than L1 success rates tests the
stratification reference (Section~\ref{sec:behavior-analysis}), but selecting
the intermediate condition on the same evaluation used for the difficulty
analysis remains a source of selection bias.  One model judge supplies both the
semantic audits and the trajectory annotations, so replicated prompts bound
prompt sensitivity but not judge independence.

Broader validation should extend the comparison to other model families and
domains, disentangle semantics from length, and study compilers that construct
useful contexts without reference solutions.  The ordering across scales
motivates studying whether useful context changes as the same student improves.
In-domain scores do not improve monotonically across checkpoints, so multi-round
OPSD with online context selection and stability-aware optimization is a natural
next setting.  The hierarchy introduced here provides a controlled interface
for such work.

\section{Diagnostic Analyses}

\subsection{Distributional Discrepancy}
\label{app:distributional}

We deterministically sample 600 aligned training problems with seed 20260804.
For each scale, the unadapted model produces two continuations per problem
from the exact OPSD student prompt with thinking disabled, using temperature
1.1, top-$p$ 0.95, top-$k$ 20, and at most 1,024 tokens.  Each continuation
is held fixed while the same base checkpoint scores it under the student
prompt, a teacher prompt with thinking enabled but no hint, and the exact L1--L5
teacher packages.  Logits are normalized over the full vocabulary at
temperature 1.1; prompts are not truncated.  We average
token-level measurements within each rollout and divide each continuation into
four equal bins by relative position.  The analyses of whole trajectories and quartiles
use all 3,600 scored rollouts, with 1,200 per scale.  Bootstrap intervals in
the diagnostic figures resample problems together with their associated
rollouts, and therefore quantify evaluation uncertainty conditional on the
evaluated runs rather than optimization variability.

\phantomsection
\label{app:answer-aligned-geometry}
For the follow-up temporal diagnostic, we locate the last
\texttt{\textbackslash boxed} marker by exact decode--re-encode token offsets.
We define a pre-answer window from up to 32 tokens immediately before the
marker and an answer-onward window from the marker through at most 32 tokens,
requiring at least eight tokens in each window.  This yields 610, 636, and 630
valid rollouts at 1.7B, 4B, and 8B, spanning 358, 371, and 364 problems.  The
1,024-token generation limit truncates 505, 488, and 524 of the 1,200
rollouts at the respective scales (40.7--43.7\%); most truncated trajectories
never reach a boxed answer and are therefore absent from the aligned subset.
The subset consequently favors completed trajectories containing a boxed answer.

\begin{figure*}[t]
\centering
\includegraphics[width=0.96\textwidth]{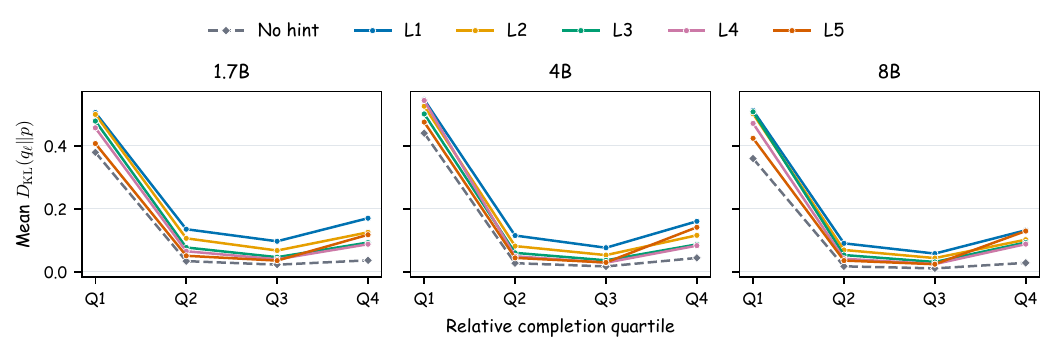}
\caption{Temporal geometry at each scale, using rollouts from the exact OPSD student
prompt.  Curves report mean unclipped full-vocabulary teacher-to-student KL
within quartiles of relative completion position, using all 1,200 rollouts at
each scale; Q4 is the final quarter of each rollout, not the answer region, and
this analysis therefore avoids the boxed-answer filtering above without
eliminating truncation bias.  The no-hint reference uses the thinking-enabled
teacher interface without privileged content.  L5 has the smallest mean
displacement among L1--L5 through Q2, remains among the smallest in Q3, and has
the largest rebound from Q3 to Q4 at every scale, both features that
means over whole trajectories hide.}
\label{fig:supervision-quartiles}
\end{figure*}

As an outcome-conditioned check, we repeat the paired answer-onward contrast
on the 421, 510, and 499 valid rollouts whose boxed answer matches the
reference under a normalized answer-string heuristic.  The resulting
$\mathrm{L5}-\mathrm{L1}$ contrasts are $-0.193$, $+0.030$, and $+0.188$,
preserving the scale-dependent sign change.  This normalized answer-string
check confirms that the sign change persists within apparently correct
rollouts; it remains an outcome-composition diagnostic.
Figure~\ref{fig:supervision-quartiles} reports the corresponding
quartile-resolved displacement curves.

\subsection{Item-Level and Behavioral Analysis}
\label{app:behavior-appendix}
\label{app:judge-protocol}

The frozen manifest for behavioral analysis contains 120 runs evaluated at step 200 and
55,134 problem--condition records, with eight shards per run and no failures
in the deterministic artifact analysis.  The manifest predates the later
addition of the 4B/8B in-domain Base evaluations, which are used in the main
score tables but are not required by the behavioral contrasts between L1 and
intermediate contexts.  For sampled-answer tasks, per-generation accuracy
(\emph{reliability}) is complemented by coverage (at least one correct
generation) and majority-vote consensus; MT-AIME is grouped by source problem
for cross-language analysis, and for ZebraLogic the gap between cell and
exact-puzzle accuracy measures whether locally correct constraints close into a complete grid.
Response length uses a tokenizer-free word/punctuation proxy and is compared
only within a benchmark.

For the model-judged behaviors, the frozen manifest requests four generations
for each of 30 matched AIME24 problems and 60 matched GPQA problems per scale
and available condition.  After structural validation, behavior rates use up to
four valid generations per problem (5,997 of 6,000 requested annotations
overall).  Trajectories longer than 24,000 characters retain their beginning and
end.  The manifest stratifies problems by the reference present when it was
frozen: four of the six scale--benchmark groups use Base, while 4B and 8B AIME24
use L1 because their Base artifacts were added later.  This stratification only
balances problem selection; the reported behavior rates pool across difficulty
bins.  The judge additionally labels method family, method switching, and
answer-before-justification beyond the fields discussed in the main text.

The trajectory judge used in Section~\ref{sec:behavior-analysis} follows one
zero-temperature system prompt:
\begin{promptbox}
You are a careful annotator of mathematical and logical reasoning
trajectories. Judge only the behavior visible in the trajectory. Do not
reward verbosity and do not infer hidden intent. Return one JSON object with
exactly these fields: \{primary\_method, secondary\_method,
method\_switch\_count, verification\_present, backtracking\_present,
explicit\_error\_correction, successful\_recovery,
answer\_before\_justification, premature\_commitment, reasoning\_complete,
brief\_evidence\}

\medskip
Definitions:\\
- A method switch requires abandoning or materially replacing an approach,
not routine algebra.\\
- successful\_recovery requires a visible mistake or rejected path followed
by a valid recovery.\\
- answer\_before\_justification means a candidate/final answer is stated
before its supporting derivation.\\
- premature\_commitment means the trajectory commits early and fails to
adequately test the commitment.\\
- Normalize primary\_method labels, e.g. algebra, geometry, counting,
number\_theory, case\_analysis, contradiction, constraint\_propagation,
elimination, scientific\_knowledge, estimation, other.
\end{promptbox}
The user message contains the problem, the ground-truth answer, the model's
extracted final answer, and the trajectory (truncated as described above).  The judge runs at temperature zero
with a 500-token response budget, and structurally invalid JSON responses
are retried.

\begin{figure}[!htbp]
\centering
\includegraphics[width=\textwidth]{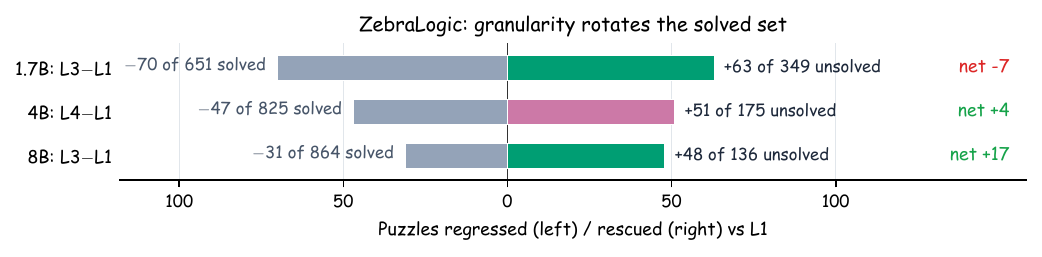}
\caption{Whether rotation pays depends on capacity.  For each scale, the bars
count ZebraLogic puzzles the strongest intermediate condition rescues from
L1-unsolved (right, level color) and gives back from L1-solved (left, gray)
at step 200.  The rescue rate rises with scale (18\% to 35\% of L1-unsolved
puzzles) while regressions shrink, turning a net loss at 1.7B into a net gain
of +17 puzzles at 8B.}
\label{fig:solved-set-rotation}
\end{figure}

\begin{figure*}[!htbp]
\centering
\includegraphics[width=0.98\textwidth]{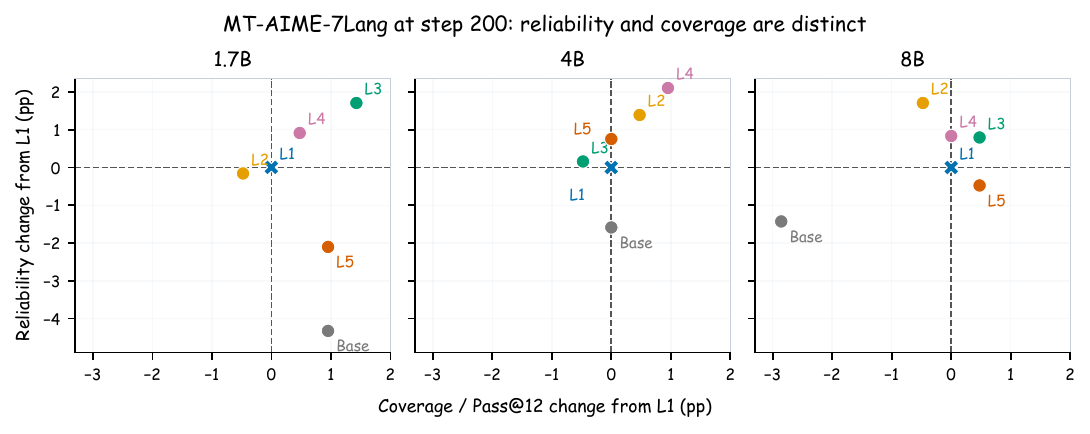}
\caption{MT-AIME-7Lang reliability and coverage at the fixed
checkpoint at step 200.  Both axes report percentage-point changes from the same-scale L1
condition, so the three panels share a directly comparable coordinate system.
The upper-right quadrant improves both per-generation reliability and
Pass@12 coverage; points above but left of the origin improve reliability
without expanding the set of problems reached by twelve samples.}
\label{fig:mt-coverage-reliability}
\end{figure*}

\begin{figure}[!htbp]
\centering
\begin{minipage}[c]{0.50\textwidth}
\centering
\includegraphics[width=0.9\linewidth]{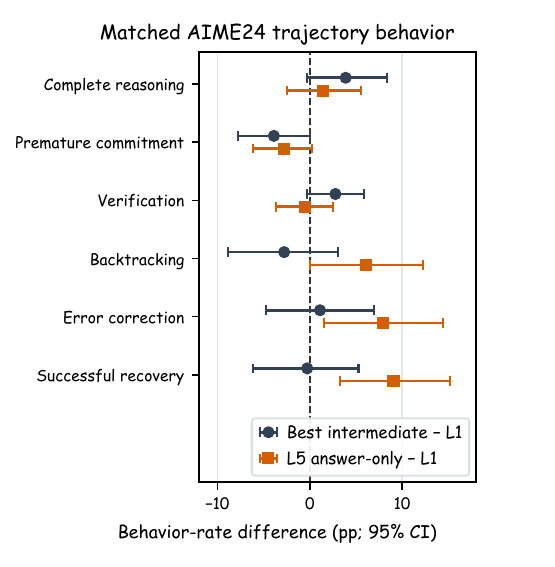}
\end{minipage}\hfill
\begin{minipage}[c]{0.46\textwidth}
\caption{Matched AIME24 differences in model-judged behavior rates relative to L1
at step 200, pooled across scales, for the highest-scoring intermediate
context and for answer-only L5 (95\% bootstrap intervals over problems).  Pooling
buys interval estimates at the cost of resolution by scale; the underlying
rates for each scale appear in Table~\ref{tab:judge-behavior}.}
\label{fig:judge-behavior}
\end{minipage}
\end{figure}

These floats support the item-level and behavioral analyses of
Section~\ref{sec:behavior-analysis}.  Table~\ref{tab:difficulty-redistribution}
gives the difficulty-conditioned changes; because its buckets are defined by
L1's own per-item success rate, part of the easy-bucket decline and the
unsolved-bucket recovery is expected from regression to the mean, which is why
Section~\ref{sec:behavior-analysis} repeats the grouping with Base success
rates.  Figure~\ref{fig:solved-set-rotation} gives the ZebraLogic
rescue-versus-regression counts, and
Figure~\ref{fig:mt-coverage-reliability} separates MT-AIME
per-generation reliability from Pass@12 coverage.
Table~\ref{tab:judge-behavior} reports the per-scale judged behavior rates and
Figure~\ref{fig:judge-behavior} the pooled differences with bootstrap
intervals.

\begin{table*}[!htbp]
\caption{Difficulty-conditioned redistribution relative to L1 at step 200.
Difficulty buckets use L1's per-item success rate with the thresholds defined
in Section~\ref{sec:behavior-analysis}.
New solve is the fraction of L1-unsolved evaluation instances for which the
comparison condition produces at least one correct sample.  Medium and easy
report changes in per-generation correctness.  MT-AIME difficulty rows treat
each translated instance separately.}
\label{tab:difficulty-redistribution}
\centering
\small
\begin{tabular*}{\textwidth}{@{\extracolsep{\fill}}lllcccc@{}}
\toprule
Model & Benchmark & Contrast & Overall $\Delta$ & New solve & Medium $\Delta$ & Easy $\Delta$ \\
\midrule
1.7B & MT-AIME & \LIII\,\mn\,\LI & +1.71 & 12.3 & +5.65 & $-4.68$ \\
4B   & MT-AIME & \LIV\,\mn\,\LI & +2.10 & 16.7 & +8.87 & +0.68 \\
8B   & MT-AIME & \LII\,\mn\,\LI & +1.71 & 21.4 & +15.79 & $-0.11$ \\
1.7B & GPQA & \LIII\,\mn\,\LI & +0.81 & 24.2 & +0.67 & $-3.70$ \\
4B   & GPQA & \LII\,\mn\,\LI & +1.21 & 25.0 & +2.63 & $-1.12$ \\
8B   & GPQA & \LIII\,\mn\,\LI & +1.41 & 25.6 & +5.26 & $-0.96$ \\
1.7B & ZebraLogic & \LIII\,\mn\,\LI & $-0.70$ & 18.1 & -- & $-10.75$ \\
4B   & ZebraLogic & \LIV\,\mn\,\LI & +0.40 & 29.1 & -- & $-5.70$ \\
8B   & ZebraLogic & \LIII\,\mn\,\LI & +1.70 & 35.3 & -- & $-3.59$ \\
\bottomrule
\end{tabular*}
\end{table*}

\begin{table*}[!htbp]
\caption{Matched AIME24 behavior at step 200.  $I^*$ is the strongest
intermediate condition on the full AIME24 evaluation at step 200 (L3 for 1.7B;
L4 for 4B and 8B).  Avg@12 uses all 360 samples.  Behavior percentages use the
judge-stratified subset of 30 problems and up to four valid generations per
problem and condition.}
\label{tab:judge-behavior}
\centering
\small
\begin{tabular*}{\textwidth}{@{\extracolsep{\fill}}llcccccc@{}}
\toprule
Model & Condition & Avg@12 & Complete & Premature & Verify & Backtrack & Correction \\
\midrule
\multirow{3}{*}{1.7B}
 & \LI & 53.89 & 65.8 & 37.5 & 69.2 & 47.5 & 33.3 \\
 & $I^*{=}$\,\LIII & 55.00 & 70.0 & 31.7 & 74.2 & 50.0 & 40.0 \\
 & \LV & 53.61 & 70.0 & 32.5 & 69.2 & 60.0 & 47.5 \\
\midrule
\multirow{3}{*}{4B}
 & \LI & 73.33 & 79.2 & 19.2 & 83.3 & 36.7 & 28.3 \\
 & $I^*{=}$\,\LIV & 76.11 & 85.0 & 17.5 & 85.8 & 30.8 & 28.3 \\
 & \LV & 73.89 & 80.8 & 18.3 & 80.8 & 41.7 & 32.5 \\
\midrule
\multirow{3}{*}{8B}
 & \LI & 78.61 & 85.0 & 18.3 & 87.5 & 35.0 & 30.0 \\
 & $I^*{=}$\,\LIV & 78.33 & 86.7 & 14.2 & 88.3 & 30.0 & 26.7 \\
 & \LV & 76.67 & 83.2 & 16.0 & 88.2 & 36.1 & 35.3 \\
\bottomrule
\end{tabular*}
\end{table*}

Length does not move in one direction with the gains: selected intermediate
responses are usually 4--10\% longer than L1 on MT-AIME and GPQA but 3.7\%
shorter on 8B ZebraLogic.  A uniform length effect is therefore ruled out,
though length may still contribute on individual tasks.

One contrast case makes the cell-to-puzzle gap concrete.  To keep the choice
independent of the narrative, we
show one case from the predeclared contrast set (winner per-item success
$\ge 0.75$, loser $\le 0.25$, selected before reading any trajectory).  The
case is ZebraLogic puzzle \texttt{lgp-test-4x6-25} at 8B: four houses with
six attribute categories.  L3 solves the grid in a 33,912-character
trajectory that terminates normally; L1 produces a 67,938-character
trajectory, also terminating normally, whose final grid is wrong.  Both
models make extensive local progress; they differ in whether that progress
closes into a unique solution.

Clue 6 states ``there are two houses between the photography enthusiast and
the person who loves beach vacations,'' which in this benchmark's convention
requires positional distance three.  The L3 trajectory parses the constraint
correctly at first contact (``\emph{if photography is in house Y, then beach is
in Y+3 or Y-3}''), which collapses the search space to two placements and lets
the remaining deductions close the grid.  The L1 trajectory instead adopts a
distance-two reading, accepts photography in house~2 with beach in house~4, and
ratifies the misreading in its final self-check (``\emph{Clue 6: two houses
between photography (House 2) and beach (House 4). Yes, Houses 3 between.
[\ldots] All clues are satisfied!}'').

Because the misread constraint underdetermines the grid, the L1 trajectory
reaches its final answer with two cells still unresolved and commits by
appeal to uniqueness rather than by deduction:
\begin{promptbox}
If I assign House 1: mountain and House 3: cruise, that's one possibility.
Alternatively, House 1: cruise and House 3: mountain. Both are possible.
But since the problem requires a unique solution, and I have no more clues,
I'll proceed with one of them. Let me choose House 1: mountain and House 3:
cruise.
\end{promptbox}
The trajectory verifies each clue against its own misreading, carries the
misread constraint through to the end, and then guesses between alternatives
it has not resolved.  The failing L1 trajectory is twice
as long and contains more explicit contradiction and re-derivation episodes
(six ``contradiction'' mentions versus one).  In this example, success turns
on correct constraint representation and closure, not response length or
visible checking activity.  The selection rule and the remaining 58 contrast
cases are part of the analysis artifact in the code repository.

\section{Context Assignment}
\label{app:shared-bridge}

The 4B prompt-control runs replace every level-specific label, delimiter, and
transition (Appendix~\ref{app:bridge-prompts}) with the following text:
\begin{promptbox}
Here is additional information for this problem:\\
=== Additional Information Begin ===\\
\{hint\}\\
=== Additional Information End ===

\medskip
Use the additional information above to independently solve the problem.\\
Do not merely copy or cite it. Reason step by step, justify the necessary
steps, and put the final answer within \textbackslash boxed\{\}.
\end{promptbox}
The student prompt is unchanged and never contains the hint.  Relative to the
primary 4B runs, only the teacher-side wrapper and transition are made common;
the underlying L1--L5 contexts, 29,434 training problems, optimizer settings,
checkpoint schedule, and evaluators remain fixed.

\begin{table*}[!htbp]
\caption{Qwen3-4B shared-bridge and mixture ablations.  ID columns report the
best Avg@12 up to step 200 (selected step in parentheses); MT-AIME and GPQA
use step 200.  Mixtures assign one hint per problem.}
\label{tab:shared-bridge}
\centering
\small
\setlength{\tabcolsep}{4pt}
\begin{tabular*}{\textwidth}{@{\extracolsep{\fill}}lcccccc@{}}
\toprule
Condition & AIME24 & AIME25 & HMMT25 & ID mean & MT-AIME & GPQA \\
\midrule
Base & 75.00 & 64.44 & 41.39 & 60.28 & 65.87 & 53.69 \\
\midrule
\LI: solution & 75.83 (100) & 68.33 (50)  & 44.72 (150) & 62.96 & 66.75 & \second{55.51} \\
\LII: strategy & \second{77.22} (150) & 68.89 (150) & \best{47.50} (150) & 64.54 & \second{69.40} & 55.25 \\
\LIII: framing  & \second{77.22} (175) & 68.61 (150) & 46.11 (150) & 63.98 & 68.57 & 55.45 \\
\LIV: category & \best{78.06} (100) & \best{71.39} (200) & \second{46.39} (25) & \best{65.28} & 68.57 & 55.45 \\
\LV: answer   & 76.11 (150) & 70.00 (175) & 45.56 (175) & 63.89 & 67.34 & \best{55.61} \\
\midrule
Uniform \LI--\LIV & 76.11 (150) & 68.89 (50)  & 45.56 (150) & 63.52 & 67.74 & 55.15 \\
MidMix \LII/\LIII   & 76.67 (50)  & \second{70.83} (125) & \second{46.39} (175) & \second{64.63} & \best{69.52} & 55.20 \\
ExtremeMix \LI/\LIV & 75.83 (125) & 69.72 (100) & 46.11 (200) & 63.89 & 67.02 & 54.75 \\
\bottomrule
\end{tabular*}
\end{table*}

Each mixture contains 29,434 rows with exactly one hint per problem,
deterministically under seed 20260720: Uniform uses 7,359 L1, 7,359 L2, 7,358
L3, and 7,358 L4 rows, MidMix 14,717 L2 and 14,717 L3, ExtremeMix 14,717 L1
and 14,717 L4.  They test balanced exposure under a fixed example budget, not
concatenated contexts.  Routed variants reuse these assignments but invoke
each row's source-level wrapper and transition, so an L1 row is formatted
exactly as a primary L1 example.  Each mixture is trained once at 4B and 8B
under the primary configuration for that scale, 200 steps.  \emph{Routed}
therefore denotes deterministic prompt construction, not a learned router or
test-time selection.

Per-sample policies use Qwen3-4B, the level-specific bridges, and the training
configuration of Table~\ref{tab:training-config}; only the context assignment
varies.  The routing signal is the \emph{hint advantage}
$\Delta_\ell=\overline{\log q_\ell(y)}-\overline{\log q_\emptyset(y)}$, the
mean sampled-token log-probability advantage of a hinted teacher over a
hint-free teacher baseline $q_\emptyset$ on the same rollout.

We selected it offline on 1,200 training problems with four base-model
rollouts each.  The absolute student--teacher log-probability gap separates
levels weakly (0.306--0.325 nats) and teacher mass on the
student's top-$k$ support saturates above 0.978 even at $k{=}4$, whereas
$\Delta_\ell$ has a median between- to within-level variance ratio of 4.5
against 1.6 for that gap, and a per-problem argmax stable across rollouts for
78\% of problems.  Its per-level means ($-0.020$, $-0.027$, $-0.042$, $-0.084$
for L3, L4, L2, L1) are negative throughout, lowest for L1, and match the
fixed-level ordering.

Four policies assign one teacher context per trajectory:
(i)~\emph{learned router}: a two-layer classifier over 39 post-rollout
features, trained online every ten optimizer steps from oracle labels on the
current batch; (ii)~\emph{likelihood-advantage routing}: each step scores all
four hinted contexts plus $q_\emptyset$ (five extra no-gradient forwards) and
routes by $\arg\max_\ell \Delta_\ell$; (iii)~\emph{aligned rule}: correct
rollouts (by a boxed-answer check against the training answer) to L3 and
incorrect ones to L4, following the offline difficulty direction;
(iv)~\emph{reversed rule}: correct to L3 and incorrect to L1, on the intuition
that failure needs the strongest supervision.  The rules use no extra teacher
forwards.

\begin{table}[!htbp]
\caption{Per-sample adaptive granularity selection at 4B (Avg@12).
Per-benchmark columns and the ID mean report the best checkpoint up to step
200; ID@200 reports the mean at the fixed endpoint.  Fixed L1 and L3 repeat the
corresponding Table~\ref{tab:id-main} conditions.  No adaptive policy
exceeds the best fixed level on the peak mean, and the reversed rule falls
below fixed L1.}
\label{tab:adaptive-routing}
\centering
\small
\setlength{\tabcolsep}{4pt}
\begin{tabular}{@{}lrrrrr@{}}
\toprule
Condition & AIME24 & AIME25 & HMMT25 & ID mean & ID@200 \\
\midrule
Fixed \LI                       & 76.94 & 68.61 & 44.72 & 63.43 & 60.93 \\
Fixed \LIII (best fixed)          & \best{77.50} & 69.72 & \best{47.22} & \best{64.81} & 61.76 \\
\midrule
Learned router                 & 76.67 & \best{71.11} & 46.39 & \second{64.72} & \second{62.69} \\
Likelihood-advantage routing ($\Delta_\ell$) & 76.94 & 69.44 & 46.11 & 64.17 & \best{62.87} \\
Rule: correct$\to$\LIII, incorrect$\to$\LIV & 76.11 & 70.83 & \second{46.67} & 64.54 & 62.59 \\
Rule: correct$\to$\LIII, incorrect$\to$\LI & 75.56 & 67.78 & 45.83 & 63.06 & 61.76 \\
\bottomrule
\end{tabular}
\end{table}

The shared bridge preserves the separation between L1 and intermediate
contexts but moves the nominal leader from L3 to L4: the family-level contrast
is stable, the exact L2--L4 ranking is not.

No mixture and no per-sample policy exceeds the best fixed level on the ID
peak mean (Tables~\ref{tab:shared-bridge} and~\ref{tab:adaptive-routing}),
including likelihood-advantage routing that computes $\Delta_\ell$ exactly at
every step.

On transfer, the comparison depends on the reference: MidMix reaches 69.52 on MT-AIME, above
the best fixed level with the shared bridge (L2, 69.40) but below the best
fixed level with the primary bridges (L4, 69.56 in Table~\ref{tab:ood-main}),
and no mixture leads on GPQA.

\begin{figure*}[!htbp]
\centering
\includegraphics[width=0.98\textwidth]{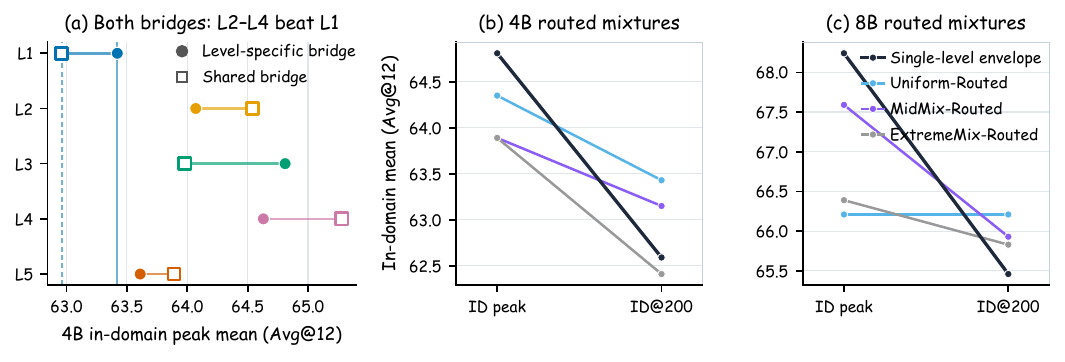}
\caption{Controls for prompt wording and mixtures.
\textbf{(a)}~Shared-bridge replication: 4B in-domain peak mean (Avg@12) for
each level under the level-specific and the level-agnostic bridge template.
L2--L4 exceed L1 under both templates, so the separation persists under common
teacher wording.
\textbf{(b,\,c)}~Routed mixtures under the level-specific bridge at 4B
and 8B, compared against the single-level envelope. Five of the six routed
mixtures exceed the best fixed-level endpoint at step 200, while none exceeds the
best fixed-level in-domain peak.}
\label{fig:bridge-mixtures}
\end{figure*}

Several policies improve the fixed endpoint at step 200: five of six routed
mixtures exceed the corresponding best fixed-level endpoint by 0.37--0.84
points while one trails by 0.18 (Table~\ref{tab:routed-mixtures};
Figure~\ref{fig:bridge-mixtures}), and the learned and likelihood-advantage
routers exceed fixed L3's endpoint by 0.93 and 1.11 points while leaving the
best observed peak unchanged.  Direction matters too: sending incorrect
rollouts to L1 rather than L4 lowers the peak mean by 1.48 points and falls
below fixed L1.  These are single runs, so they suggest reduced sensitivity to
the stopping point rather than a higher attainable peak.

\begin{table*}[!htbp]
\caption{Level-conditioned mixture controls.  ID peak averages the
benchmark-wise best Avg@12 up to step 200; ID@200 averages all three ID
benchmarks at the common terminal checkpoint.  MT-AIME and GPQA use step 200.
The single-level envelope takes the best L1--L5 result independently in each
column and may therefore select different levels.  Routed rows use the
original level-specific bridge associated with each example's assigned level.
Light-blue/bold shading marks the column maximum within scale;
light-gray/underline shading marks the second-best.}
\label{tab:routed-mixtures}
\centering
\small
\setlength{\tabcolsep}{4pt}
\begin{tabular*}{\textwidth}{@{\extracolsep{\fill}}llcccc@{}}
\toprule
Scale & Condition & ID peak & ID@200 & MT-AIME & GPQA \\
\midrule
\multirow{5}{*}{4B}
 & Base & 60.28 & -- & 65.87 & 53.69 \\
 & Single-level envelope & \best{64.81} & 62.59 & \best{69.56} & \best{55.96} \\
 & Uniform-Routed & \second{64.35} & \best{63.43} & 68.29 & 54.85 \\
 & MidMix-Routed & 63.89 & \second{63.15} & 67.70 & \second{55.30} \\
 & ExtremeMix-Routed & 63.89 & 62.41 & \second{69.13} & 55.05 \\
\midrule
\multirow{5}{*}{8B}
 & Base & 62.13 & -- & 70.48 & 60.05 \\
 & Single-level envelope & \best{68.24} & 65.46 & \best{73.61} & \best{62.68} \\
 & Uniform-Routed & 66.21 & \best{66.21} & 72.22 & \second{61.62} \\
 & MidMix-Routed & \second{67.59} & \second{65.93} & \second{72.90} & 61.01 \\
 & ExtremeMix-Routed & 66.39 & 65.83 & 72.26 & 61.11 \\
\bottomrule
\end{tabular*}
\end{table*}

%% file: refs.bib
@inproceedings{
zhao2026self,
title={Self-Distilled Reasoner: On-Policy Self-Distillation for Large Language Models},
author={Siyan Zhao and Zhihui Xie and Mengchen Liu and Jing Huang and Guan Pang and Feiyu Chen and Aditya Grover},
booktitle={Forty-third International Conference on Machine Learning},
year={2026},
url={https://openreview.net/forum?id=Jpxfof0EaS}
}

@article{qwen2025qwen3,
  title   = {Qwen3 Technical Report},
  author  = {{Qwen Team}},
  journal = {arXiv preprint arXiv:2505.09388},
  year    = {2025}
}

@article{rein2023gpqa,
  title   = {GPQA: A Graduate-Level Google-Proof Q\&A Benchmark},
  author  = {Rein, David and Hou, Betty Li and Stickland, Asa Cooper and Petty, Jackson and Pang, Richard Yuanzhe and Dirani, Julien and Michael, Julian and Bowman, Samuel R.},
  journal = {arXiv preprint arXiv:2311.12022},
  year    = {2023}
}

@article{huang2023ceval,
  title   = {C-Eval: A Multi-Level Multi-Discipline Chinese Evaluation Suite for Foundation Models},
  author  = {Huang, Yuzhen and Bai, Yuzhuo and Zhu, Zhihao and Zhang, Junlei and Zhang, Jinghan and Su, Tangjun and Liu, Junteng and Lv, Chuancheng and Zhang, Yikai and Lei, Jiayi and Fu, Yao and Sun, Maosong and He, Junxian},
  journal = {Advances in Neural Information Processing Systems},
  volume  = {36},
  year    = {2023}
}

@inproceedings{zelikman2022star,
author = {Zelikman, Eric and Wu, Yuhuai and Mu, Jesse and Goodman, Noah D.},
title = {STaR: self-taught reasoner bootstrapping reasoning with reasoning},
year = {2022},
isbn = {9781713871088},
publisher = {Curran Associates Inc.},
address = {Red Hook, NY, USA},
booktitle = {Proceedings of the 36th International Conference on Neural Information Processing Systems},
articleno = {1126},
numpages = {13},
location = {New Orleans, LA, USA},
series = {NIPS '22}
}

@inproceedings{hsieh2023distilling,
    title = "Distilling Step-by-Step! Outperforming Larger Language Models with Less Training Data and Smaller Model Sizes",
    author = "Hsieh, Cheng-Yu  and
      Li, Chun-Liang  and
      Yeh, Chih-kuan  and
      Nakhost, Hootan  and
      Fujii, Yasuhisa  and
      Ratner, Alex  and
      Krishna, Ranjay  and
      Lee, Chen-Yu  and
      Pfister, Tomas",
    editor = "Rogers, Anna  and
      Boyd-Graber, Jordan  and
      Okazaki, Naoaki",
    booktitle = "Findings of the Association for Computational Linguistics: ACL 2023",
    month = jul,
    year = "2023",
    address = "Toronto, Canada",
    publisher = "Association for Computational Linguistics",
    url = "https://aclanthology.org/2023.findings-acl.507/",
    doi = "10.18653/v1/2023.findings-acl.507",
    pages = "8003--8017"
}

@inproceedings{lightman2023verify,
  title     = {Let's Verify Step by Step},
  author    = {Lightman, Hunter and Kosaraju, Vineet and Burda, Yura and Edwards, Harri and Baker, Bowen and Lee, Teddy and Leike, Jan and Schulman, John and Sutskever, Ilya and Cobbe, Karl},
  booktitle = {International Conference on Learning Representations},
  year      = {2024}
}

@article{kaur2026rethinking,
  title         = {Rethinking On-Policy Self-Distillation for Thinking Models},
  author        = {Kaur, Simran and Ri, Narutatsu and He, Yinghui and Fowl, Liam and Arora, Sanjeev},
  journal       = {arXiv preprint arXiv:2607.05184},
  year          = {2026},
  eprint        = {2607.05184},
  archivePrefix = {arXiv},
  primaryClass  = {cs.LG},
  url           = {https://arxiv.org/abs/2607.05184}
}

@inproceedings{son2025linguistic,
  title     = {Linguistic Generalizability of Test-Time Scaling in Mathematical Reasoning},
  author    = {Son, Guijin and Hong, Jiwoo and Ko, Hyunwoo and Thorne, James},
  booktitle = {Proceedings of the 63rd Annual Meeting of the Association for Computational Linguistics (Volume 1: Long Papers)},
  pages     = {14333--14368},
  year      = {2025},
  address   = {Vienna, Austria},
  publisher = {Association for Computational Linguistics},
  doi       = {10.18653/v1/2025.acl-long.699},
  url       = {https://aclanthology.org/2025.acl-long.699/}
}

@misc{zhu2025autologi,
      title={AutoLogi: Automated Generation of Logic Puzzles for Evaluating Reasoning Abilities of Large Language Models}, 
      author={Qin Zhu and Fei Huang and Runyu Peng and Keming Lu and Bowen Yu and Qinyuan Cheng and Xipeng Qiu and Xuanjing Huang and Junyang Lin},
      year={2025},
      eprint={2502.16906},
      archivePrefix={arXiv},
      primaryClass={cs.CL},
      url={https://arxiv.org/abs/2502.16906}, 
}

@InProceedings{lin2025zebralogic,
  title = 	 {{Z}ebra{L}ogic: On the Scaling Limits of {LLM}s for Logical Reasoning},
  author =       {Lin, Bill Yuchen and Le Bras, Ronan and Richardson, Kyle and Sabharwal, Ashish and Poovendran, Radha and Clark, Peter and Choi, Yejin},
  booktitle = 	 {Proceedings of the 42nd International Conference on Machine Learning},
  pages = 	 {37889--37905},
  year = 	 {2025},
  editor = 	 {Singh, Aarti and Fazel, Maryam and Hsu, Daniel and Lacoste-Julien, Simon and Berkenkamp, Felix and Maharaj, Tegan and Wagstaff, Kiri and Zhu, Jerry},
  volume = 	 {267},
  series = 	 {Proceedings of Machine Learning Research},
  month = 	 {13--19 Jul},
  publisher =    {PMLR},
  url = 	 {https://proceedings.mlr.press/v267/lin25i.html}
}

@article{vapnik2009new,
  title={A new learning paradigm: Learning using privileged information},
  author={Vapnik, Vladimir and Vashist, Akshay},
  journal={Neural Networks},
  volume={22},
  number={5-6},
  pages={544--557},
  year={2009}
}

@inproceedings{lopezpaz2016unifying,
title     = {Unifying Distillation and Privileged Information},
author    = {Lopez-Paz, David and Bottou, L{\'e}on and Sch{\"o}lkopf, Bernhard and Vapnik, Vladimir},
booktitle = {International Conference on Learning Representations},
pages     = {1--10},
year      = {2016}
}

@article{hinton2015distilling,
  title={Distilling the knowledge in a neural network},
  author={Hinton, Geoffrey and Vinyals, Oriol and Dean, Jeff},
  journal={arXiv preprint arXiv:1503.02531},
  year={2015}
}

@inproceedings{agarwal2024onpolicy,
  title={On-policy distillation of language models: Learning from self-generated mistakes},
  author={Agarwal, Rishabh and Vieillard, Nino and Zhou, Yongchao and Stanczyk, Piotr and Ramos, Sabela and Geist, Matthieu and Bachem, Olivier},
  booktitle={International Conference on Learning Representations},
  year={2024}
}

@article{snell2022context,
  title={Learning by distilling context},
  author={Snell, Charlie and Klein, Dan and Zhong, Ruiqi},
  journal={arXiv preprint arXiv:2209.15189},
  year={2022}
}

@inproceedings{gu2024minillm,
 author = {Gu, Yuxian and Dong, Li and Wei, Furu and Huang, Minlie},
 booktitle = {International Conference on Learning Representations},
 editor = {B. Kim and Y. Yue and S. Chaudhuri and K. Fragkiadaki and M. Khan and Y. Sun},
 pages = {32694--32717},
 title = {MiniLLM: Knowledge Distillation of Large Language Models},
 url = {https://proceedings.iclr.cc/paper_files/paper/2024/file/8ac015d409635f196f9e3e9dcfb9a94e-Paper-Conference.pdf},
 volume = {2024},
 year = {2024}
}

@article{ye2026onpolicycontext,
  title         = {On-Policy Context Distillation for Language Models},
  author        = {Ye, Tianzhu and Dong, Li and Wu, Xun and Huang, Shaohan and Wei, Furu},
  journal       = {arXiv preprint arXiv:2602.12275},
  year          = {2026},
  eprint        = {2602.12275},
  archivePrefix = {arXiv},
  primaryClass  = {cs.CL},
  url           = {https://arxiv.org/abs/2602.12275}
}

@inproceedings{
penaloza2026privileged,
title={Privileged Information Distillation for Language Models},
author={Emiliano Penaloza and Dheeraj Vattikonda and Nicolas Gontier and Alexandre Lacoste and Laurent Charlin and Massimo Caccia},
booktitle={Forty-third International Conference on Machine Learning},
year={2026},
url={https://openreview.net/forum?id=ebZcMQImhG}
}

@article{li2026rethinkingopd,
  title         = {Rethinking On-Policy Distillation of Large Language Models: Phenomenology, Mechanism, and Recipe},
  author        = {Li, Yaxuan and Zuo, Yuxin and He, Bingxiang and Zhang, Jinqian and Xiao, Chaojun and Qian, Cheng and Yu, Tianyu and Gao, Huan-ang and Yang, Wenkai and Liu, Zhiyuan and Ding, Ning},
  journal       = {arXiv preprint arXiv:2604.13016},
  year          = {2026},
  eprint        = {2604.13016},
  archivePrefix = {arXiv},
  primaryClass  = {cs.CL},
  url           = {https://arxiv.org/abs/2604.13016}
}

@inproceedings{
jin2026entropyaware,
title={Entropy-Aware On-Policy Distillation of Language Models},
author={Woogyeol Jin and Taywon Min and Yongjin Yang and Dennis Wei and Yi Zhou and Swanand Ravindra Kadhe and Nathalie Baracaldo and Kimin Lee},
booktitle={Forty-third International Conference on Machine Learning},
year={2026},
url={https://openreview.net/forum?id=J5i09faOOf}
}

@article{luo2026demystifying,
  title         = {Demystifying {OPD}: Length Inflation and Stabilization Strategies for Large Language Models},
  author        = {Luo, Feng and Chuang, Yu-Neng and Wang, Guanchu and Xu, Zicheng and Han, Xiaotian and Zhang, Tianyi and Braverman, Vladimir},
  journal       = {arXiv preprint arXiv:2604.08527},
  year          = {2026},
  eprint        = {2604.08527},
  archivePrefix = {arXiv},
  primaryClass  = {cs.CL},
  url           = {https://arxiv.org/abs/2604.08527}
}

@ARTICLE{sang2026reasoningcompression,
       author = {{Sang}, Hejian and {Xu}, Yuanda and {Zhou}, Zhengze and {He}, Ran and {Wang}, Zhipeng and {Sun}, Jiachen},
        title = "{CRISP: Compressed Reasoning via Iterative Self-Policy Distillation}",
      journal = {arXiv e-prints},
         year = 2026,
        month = mar,
          eid = {arXiv:2603.05433},
        pages = {arXiv:2603.05433},
          doi = {10.48550/arXiv.2603.05433},
archivePrefix = {arXiv},
       eprint = {2603.05433},
 primaryClass = {stat.ML},
       adsurl = {https://ui.adsabs.harvard.edu/abs/2026arXiv260305433S}
}

@inproceedings{
sdpo,
title={Reinforcement Learning via Self-Distillation},
author={Jonas H{\"u}botter and Frederike L{\"u}beck and Lejs Deen Behric and Anton Baumann and Marco Bagatella and Daniel Marta and Ido Hakimi and Idan Shenfeld and Thomas Kleine Buening and Carlos Guestrin and Andreas Krause},
booktitle={Forty-third International Conference on Machine Learning},
year={2026},
url={https://openreview.net/forum?id=QkfkxyRizZ}
}

@inproceedings{
sdft,
title={Self-Distillation Enables Continual Learning},
author={Idan Shenfeld and Mehul Damani and Jonas H{\"u}botter and Pulkit Agrawal},
booktitle={Forty-third International Conference on Machine Learning},
year={2026},
url={https://openreview.net/forum?id=qA6FgH0nnZ}
}

@misc{rlsd,
      title={Self-Distilled RLVR}, 
      author={Chenxu Yang and Chuanyu Qin and Qingyi Si and Minghui Chen and Naibin Gu and Dingyu Yao and Zheng Lin and Weiping Wang and Jiaqi Wang and Nan Duan},
      year={2026},
      eprint={2604.03128},
      archivePrefix={arXiv},
      primaryClass={cs.LG},
      url={https://arxiv.org/abs/2604.03128}, 
}

@misc{dopd,
      title={DOPD: Dual On-policy Distillation}, 
      author={Xinlei Yu and Gen Li and Qingyi Si and Guibin Zhang and Yuqi Xu and Congcong Wang and Shuai Dong and Kaiwen Tuo and Xiangyu Zeng and Kaituo Feng and Qunzhong Wang and Yang Shi and Xiaobin Hu and Xiangyu Yue and Jiaqi Wang and Shuicheng Yan},
      year={2026},
      eprint={2606.30626},
      archivePrefix={arXiv},
      primaryClass={cs.AI},
      url={https://arxiv.org/abs/2606.30626}, 
}

@misc{vicur,
      title={ViCuR: Visual Cues as Recoverable Privilege for Multimodal On-Policy Distillation}, 
      author={Kanghui Tian and Siyuan Liu and Ziang Yan and Sheng Xia and Shuai Dong and Yi Wang},
      year={2026},
      eprint={2606.05718},
      archivePrefix={arXiv},
      primaryClass={cs.CV},
      url={https://arxiv.org/abs/2606.05718}, 
}

@article{lu2025onpolicy-thinkinglab,
  author = {Kevin Lu and {Thinking Machines Lab}},
  title = {On-Policy Distillation},
  journal = {Thinking Machines Lab: Connectionism},
  year = {2025},
  note = {https://thinkingmachines.ai/blog/on-policy-distillation},
  doi = {10.64434/tml.20251026},
}

@misc{exopd,
      title={Learning beyond Teacher: Generalized On-Policy Distillation with Reward Extrapolation}, 
      author={Wenkai Yang and Weijie Liu and Ruobing Xie and Kai Yang and Saiyong Yang and Yankai Lin},
      year={2026},
      eprint={2602.12125},
      archivePrefix={arXiv},
      primaryClass={cs.LG},
      url={https://arxiv.org/abs/2602.12125}, 
}

@article{uniopd,
  title   = {{Uni-OPD}: Unifying On-Policy Distillation with a Dual-Perspective Recipe},
  author  = {Hou, Wenjin and Peng, Shangpin and Wang, Weinong and Ruan, Zheng and Zhang, Yue and Zhou, Zhenglin and Gao, Mingqi and Chen, Yifei and Wang, Kaiqi and Yang, Hongming and Zhang, Chengquan and Tian, Zhuotao and Hu, Han and Yang, Yi and Wu, Fei and Fan, Hehe},
  journal = {arXiv preprint arXiv:2605.03677},
  year    = {2026}
}

@inproceedings{
hu2021loralowrankadaptationlarge,
title={Lo{RA}: Low-Rank Adaptation of Large Language Models},
author={Edward J Hu and yelong shen and Phillip Wallis and Zeyuan Allen-Zhu and Yuanzhi Li and Shean Wang and Lu Wang and Weizhu Chen},
booktitle={International Conference on Learning Representations},
year={2022},
url={https://openreview.net/forum?id=nZeVKeeFYf9}
}

@article{harne2026privileged,
  title   = {Privileged, but Biased: How PI-Conditioned Teachers Break Self-Distillation},
  author  = {Harne, Sarthak and Karkar, Chinmay and Pandya, Yash and Awadallah, Ahmed and Nambi, Akshay},
  journal = {arXiv preprint arXiv:2608.04794},
  year    = {2026}
}

@article{li2026uopsd,
  title   = {On-Policy Self-Distillation without Any Supervision},
  author  = {Li, Yijiang and Wang, Bingyang and Liang, Yijun and Tian, Yunjie and Fu, Di and Vasconcelos, Nuno},
  journal = {arXiv preprint arXiv:2608.06296},
  year    = {2026}
}

@inproceedings{
guha2025openthoughtsdatarecipesreasoning,
title={OpenThoughts: Data Recipes for Reasoning Models},
author={Etash Kumar Guha and Ryan Marten and Sedrick Keh and Negin Raoof and Georgios Smyrnis and Hritik Bansal and Marianna Nezhurina and Jean Mercat and Trung Vu and Zayne Rea Sprague and Ashima Suvarna and Benjamin Feuer and Leon Liangyu Chen and Zaid Khan and Eric Frankel and Sachin Grover and Caroline Choi and Niklas Muennighoff and Shiye Su and Wanjia Zhao and John Yang and Shreyas Pimpalgaonkar and Kartik sharma and Charlie Cheng-Jie Ji and Yichuan Deng and Sarah M Pratt and Vivek Ramanujan and Jon Saad-Falcon and Stutee Acharya and Jeffrey Li and Achal Dave and Alon Albalak and Kushal Arora and Blake Wulfe and Chinmay Hegde and Greg Durrett and Sewoong Oh and Mohit Bansal and Saadia Gabriel and Aditya Grover and Kai-Wei Chang and Vaishaal Shankar and Aaron Gokaslan and Mike A Merrill and Tatsunori Hashimoto and Yejin Choi and Jenia Jitsev and Reinhard Heckel and Maheswaran Sathiamoorthy and Alex Dimakis and Ludwig Schmidt},
booktitle={The Fourteenth International Conference on Learning Representations},
year={2026},
url={https://openreview.net/forum?id=7xjoTuaNmN}
}

@article{zhang2026illusion,
  title         = {The Illusion of Certainty: Decoupling Capability and Calibration in On-Policy Distillation},
  author        = {Zhang, Jiaxin and Peng, Xiangyu and Chen, Qinglin and Ye, Qinyuan and Xiong, Caiming and Wu, Chien-Sheng},
  journal       = {arXiv preprint arXiv:2604.16830},
  year          = {2026},
  eprint        = {2604.16830},
  archivePrefix = {arXiv},
  primaryClass  = {cs.LG},
  url           = {https://arxiv.org/abs/2604.16830}
}

@misc{wu2026regft,
      title={Learn Hard Problems During RL with Reference Guided Fine-tuning}, 
      author={Yangzhen Wu and Shanda Li and Zixin Wen and Xin Zhou and Ameet Talwalkar and Yiming Yang and Wenhao Huang and Tianle Cai},
      year={2026},
      eprint={2603.01223},
      archivePrefix={arXiv},
      primaryClass={cs.LG},
      url={https://arxiv.org/abs/2603.01223}, 
}

@article{ichihara2026privileged,
  title   = {Privileged Solutions or Context-Induced Teacher Behavior? Dissecting On-Policy Self-Distillation},
  author  = {Ichihara, Yuki and Iwase, Naoto and Quamar, Mohammad Atif and Komiyama, Junpei},
  journal = {arXiv preprint arXiv:2608.09228},
  year    = {2026}
}

@article{han2026adaptive,
  title   = {Adaptive Teacher Exposure for Self-Distillation in LLM Reasoning},
  author  = {Han, Zihao and Zhang, Tiangang and Wang, Huaibin and Sun, Yilun},
  journal = {arXiv preprint arXiv:2605.11458},
  year    = {2026}
}

@misc{shrestha2026rethinking,
      title={Rethinking Privileged Information in On-Policy Self-Distillation}, 
      author={Samyak Shrestha and Alexander Tessier},
      year={2026},
      eprint={2608.18271},
      archivePrefix={arXiv},
      primaryClass={cs.LG},
      url={https://arxiv.org/abs/2608.18271}, 
}

@article{armandpour2026unmasking,
  title={Unmasking On-Policy Distillation: Where It Helps, Where It Hurts, and Why},
  author={Mohammadreza Armandpour and Fatih Ilhan and David Harrison and Ajay Jaiswal and Duc Nam Hoang and Fartash Faghri and Yizhe Zhang and Minsik Cho and Mehrdad Farajtabar},
  journal={ArXiv},
  year={2026},
  volume={abs/2605.10889},
  url={https://api.semanticscholar.org/CorpusID:288256261}
}

@misc{qwen3.5,
    title  = {{Qwen3.5}: Towards Native Multimodal Agents},
    author = {{Qwen Team}},
    month  = {February},
    year   = {2026},
    url    = {https://qwen.ai/blog?id=qwen3.5}
}

@misc{zhao2026problemspace,
      title={Is More Privileged Information Better? From Solution Traces to Problem-Solving Structure in Self-Distilled Reasoning}, 
      author={Xuyang Zhao and Liting Zhang and Zichen Xu and Zhihu Wang and Xu Caiyue and Shiwan Zhao and Qicheng Li},
      year={2026},
      eprint={2608.01589},
      archivePrefix={arXiv},
      primaryClass={cs.AI},
      url={https://arxiv.org/abs/2608.01589}, 
}

@misc{yuan2026visionopdlearningfinedetails,
      title={Vision-OPD: Learning to See Fine Details for Multimodal LLMs via On-Policy Self-Distillation}, 
      author={Qianhao Yuan and Jie Lou and Xing Yu and Hongyu Lin and Le Sun and Xianpei Han and Yaojie Lu},
      year={2026},
      eprint={2605.18740},
      archivePrefix={arXiv},
      primaryClass={cs.CV},
      url={https://arxiv.org/abs/2605.18740}, 
}
